\documentclass[final,12pt]{colt2021} %

\usepackage{xcolor}
\usepackage{bbm}
\usepackage{dsfont}
\usepackage{pgfplots}
\usepackage{xstring}
\usepackage{trimspaces}
\usepackage{thmtools}
\usepackage{thm-restate}

\usepackage{enumitem}
\setlist{leftmargin=\parindent,topsep=0.25em,partopsep=0.25em,parsep=0.25em,itemsep=0.25em}

\DeclareMathOperator{\dif}{d \!}

\DeclareMathOperator*{\poly}{poly}

\newcommand{\EE}[2][]{\mathop{\mathbb{E}}\displaylimits_{#1}\left[ #2 \right]}
\newcommand{\EEl}[2][]{\mathop{\mathbb{E}}\displaylimits_{#1}[ #2 ]}
\newcommand{\indicator}[1]{\mathds{1}\left\{ #1 \right\}}
\newcommand{\indicatorl}[1]{\mathds{1}\{ #1 \}}

\newcommand{\pr}[2][]{\mathop{\operatorname{Pr}}\displaylimits_{#1}\left[ #2 \right]}
\newcommand{\prl}[2][]{\mathop{\operatorname{Pr}}\displaylimits_{#1}[ #2 ]}

\newcommand{\norm}[2][]{\left\| #2 \right\|_{#1}}
\newcommand{\norml}[2][]{\| #2 \|_{#1}}

\newcommand{\innerprod}[2]{\left\langle #1, #2 \right\rangle}
\newcommand{\innerprodl}[2]{\langle #1, #2 \rangle}
\newcommand{\innerproduct}[2]{\left\langle #1, #2 \right\rangle}
\newcommand{\sign}[1]{\operatorname{sign}\left( #1 \right)}
\newcommand{\signl}[1]{\operatorname{sign}( #1 )}
\newcommand{\relu}{\sigma_{\operatorname{ReLU}}}

\newcommand{\floor}[1]{\left\lfloor #1 \right\rfloor}
\newcommand{\floorl}[1]{\lfloor #1 \rfloor}
\newcommand{\ceil}[1]{\left\lceil #1 \right\rceil}
\newcommand{\ceill}[1]{\lceil #1 \rceil}
\newcommand{\abs}[1]{\left| #1 \right|}
\newcommand{\absl}[1]{| #1 |}
\renewcommand{\set}[1]{\left\{ #1 \right\}}
\newcommand{\setl}[1]{\{ #1 \}}
\newcommand{\paren}[1]{\left( #1 \right)}
\newcommand{\parenl}[1]{( #1 )}
\newcommand{\bracket}[1]{\left[#1\right]}

\newcommand{\eps}{\epsilon}

\newcommand{\N}{\mathbb{N}}
\newcommand{\R}{\mathbb{R}}
\newcommand{\Z}{\mathbb{Z}}
\newcommand{\Circle}{\mathbb{T}}

\newcommand{\flip}{\{-1, 1\}}
\newcommand{\bit}{\{0, 1\}}
\newcommand{\half}{\frac{1}{2}}
\newcommand{\halfl}{\fracl{1}{2}}

\newcommand\bb{{\mathbf{b}}}
\newcommand\bbi[1]{{\mathbf{b}^{( #1 )}}}
\newcommand\bg{{\mathbf{g}}}
\newcommand\gi[1]{{{g}^{( #1 )}}}
\newcommand\bgi[1]{{\mathbf{g}^{( #1 )}}}

\newcommand\bhi[1]{{\mathbf{h}^{( #1 )}}}
\newcommand\bu{{\mathbf{u}}}
\newcommand\bd{{\mathbf{d}}}

\newcommand\bw{{\mathbf{w}}}

\newcommand\bwi[1]{{\mathbf{w}^{( #1 )}}}

\newcommand\bx{{\mathbf{x}}}
\newcommand\by{{\mathbf{y}}}
\newcommand\bK{{\mathbf{K}}}

\newcommand\Normal{\operatorname{\mathcal{N}}}
\newcommand\MNormal{\Normal(0, I_d)}
\newcommand\vol{\operatorname{vol}}

\newcommand\minwidth[5]{\operatorname{MinWidth}_{#1, #2, #3, #4, #5}}

\newcommand\sobolev[2]{H^{#1}(#2)}
\newcommand\fnmeas[1]{L_2(#1)}
\newcommand\sobnorm[3]{\norm[\sobolev{#1}{#2}]{#3}}
\newcommand\ipmeas[3]{\innerprod{#1}{#2}_{#3}}
\newcommand\ipmeasl[3]{\innerprodl{#1}{#2}_{#3}}
\newcommand\intvl{[-1, 1]}
\newcommand\intv{[0, 1]}
\newcommand\Span[1]{\operatorname{Span}\paren{#1}}
\newcommand\Spanl[1]{\operatorname{Span}(#1)}
\newcommand\betam{\beta_{\max}}
\newcommand\sph{\mathbb{S}^{d-1}}

\newcommand{\ignore}[2][]{%
  \IfEqCase{#1}{%
    {}{}%
    {true}{}%
    {false}{#2}%
  }[\PackageError{ignore}{Undefined option to ignore: #1}{}]%
}
\newcommand{\bias}{\operatorname{bias}}
\newcommand{\weights}{\operatorname{weights}}
\newcommand{\dprod}{\mathcal{D}}
\newcommand{\dprodk}{\mathcal{D}_k}
\newcommand{\dbias}{\mathcal{D}_{\bias}}
\newcommand{\dweight}{\mathcal{D}_{\weights}}
\newcommand{\dweightk}{\mathcal{D}_{\weights, k}}

\newcommand{\supp}{\operatorname{supp}}

\renewcommand\vec\orgvec

\newcommand{\lip}{\operatorname{Lip}}

\newcommand{\pderiv}[2]{\frac{\partial #1}{\partial #2}}
\newcommand{\pderivl}[2]{\partial #1/\partial #2}

\newcommand{\dderiv}[2]{D^{(#1)} #2}

\newcommand{\spangl}{\Spanl{\bgi{j}}_{j=1}^r}

\newcommand{\sinset}{\mathcal{K}_{\sin}}
\newcommand{\cosset}{\mathcal{K}_{\cos}}

\newcommand{\fracl}[2]{#1 / #2}

\renewenvironment{proof}[1][Proof]%
{%
 \par\noindent{\bfseries\upshape #1.\ }%
}%
{\jmlrQED}

\newenvironment{proofnoqed}[1][Proof]%
{%
 \par\noindent{\bfseries\upshape #1.\ }%
}%
{}

\newcommand\qedhere{\tag*{\jmlrQED}}

\usepackage{times}
\usepackage{comment}
\usepackage{thmtools} 
\usepackage{thm-restate}
\usepackage{tikz}
\usepackage{pgfplots}
\usetikzlibrary{positioning,fit}

\title[On the Approximation Power of Two-Layer Networks of Random ReLUs]{On the Approximation Power of\\ Two-Layer Networks of Random ReLUs\ignore{Depth Separation: a Lipschitz Sensitivity Perspective}}

\coltauthor{\Name{Daniel Hsu} \Email{djhsu@cs.columbia.edu}\\
\Name{Clayton Sanford} \Email{clayton@cs.columbia.edu }\\
\Name{Rocco A. Servedio} \Email{rocco@cs.columbia.edu }\\
\Name{Emmanouil V. {Vlatakis Gkaragkounis}} \Email{emvlatakis@cs.columbia.edu }\\
\addr Columbia University}

\begin{document}

\maketitle

\begin{abstract}%
This paper considers the following question:  how well can depth-two ReLU networks with randomly initialized bottom-level weights represent smooth functions? 
We give near-matching upper- and lower-bounds for $L_2$-approximation in terms of the Lipschitz constant, the desired accuracy, and the dimension of the problem, as well as similar results in terms of Sobolev norms. Our positive results employ tools from harmonic analysis and ridgelet representation theory, while our lower-bounds are based on (robust versions of) dimensionality arguments.

\end{abstract}


\section{Introduction}
\subsection{Background and motivation} \label{sec:background}

%

Celebrated results of \cite{cyb89}, \cite{funahashi1989approximate}, and \cite{hsw89} establish the universality of depth-2 neural networks by showing that any continuous function on $\R^d$ can be approximated by a neural network with a single hidden layer.
However, these results offer no upper-bound (e.g., in terms of $d$) on the width (number of bottom-level gates) required, leaving unanswered many natural questions about the approximation power of neural networks, including:
\begin{itemize}
	\item Which functions can be approximated by two-layer neural networks of subexponential width?
	\item Can tradeoffs be achieved between depth and width for neural network function approximation?
	\item Given the practical importance of random weight initialization, what are the representational capabilities of neural networks with some randomly drawn weights (say, at the bottom level)?
\end{itemize}

The first two questions above have been studied intensely in the approximation-theoretic and depth-separation literature; this paper focuses on the third question. 
Random weight initializations play an important role in training neural networks in practice, and are also of theoretical interest; as we discuss later in this introduction, they have been well studied as a way of understanding different aspects of approximation and generalization.

In this work, we study the representational ability of depth-2 random bottom-layer (RBL) ReLU  networks. 
Such a network is equivalent to a linear combination of rectified linear units (ReLUs), where the weight vector and bias of each ReLU are randomly and independently chosen from a fixed distribution, but the top-level combining weights of the ReLUs are allowed to be arbitrary (we give precise definitions in Section \ref{ssec-rbl}).
This particular setting is of interest because, as discussed later, a number of papers have given approximation-theoretic results in this regime.
We choose the ReLU activation due to its popularity in both theory and practice; we expect that the results of our paper could be generalized to a range of other activation functions.

Our main goal is to understand the abilities and limitations of depth-2 RBL ReLU networks for approximating smooth functions of various types.
We focus on smooth functions both because they are a natural class of functions to consider, and because non-smooth functions have been shown to be difficult to approximate by various types of neural networks.
Indeed, several authors (e.g.,~\cite{tel16} and \cite{dan17-depth})
have established lower-bounds on the width of neural networks that approximate certain non-smooth functions by taking advantage of the fact that such functions can be highly oscillatory (have many ``bumps") and can require many gates to approximate each ``bump.'' 

Our chief focus is on functions over the $d$-dimensional solid cube $\intvl^d$ (though we also consider functions over $d$-dimensional Gaussian space in Appendix~\ref{asec-gaussian}) whose smoothness is measured in two different ways.
Our main results are about approximating functions on $\intvl^d$ with bounded \emph{Lipschitz constants}; in Appendix~\ref{asec:sobolev}, we also consider functions on $\intvl^d$ (satisfying certain periodicity conditions) with bounded \emph{Sobolev norms}.

\subsection{Our results}

The main contributions of this work are to pose and answer the following question: 

\begin{quote}
\emph{What is the minimum number of random ReLU features required so that (with high probability) there exists some linear combination of those features that closely approximates any sufficiently smooth function?}
\end{quote}

This minimum number of random ReLU features is equivalent to the minimum width required for a depth-2 RBL ReLU network to approximate the smooth function in question.
We give full details about our setting in Section \ref{ssec-rbl}, and here only touch on some of the main aspects:
\begin{itemize}
	
	\item ``Random ReLU features'' are functions from $\R^d$ to $\R$ that are drawn independently from some fixed distribution.
	 These take the form $x \mapsto \relu(\innerprodl{\bw}{x} + \bb)$ where $\relu(z) := \max(z, 0)$ and $\bw$ and $\bb$ are random variables taking  values in $\sph$ and $\R$ respectively.
	
	\item Our notion of ``close approximation'' refers to the $L_2$ distance between functions with respect to the uniform distribution on the solid cube; we say that $f$ is an $\eps$-approximator for $g$ if $\norm[\intvl^d]{f - g} \leq \eps$. 
	In Appendix \ref{asec-gaussian}, we sketch how analyses similar to our analysis over $\intvl^d$ can be used to study approximation with respect to the Gaussian measure over $\R^d.$
	
	\item As mentioned above, we chiefly measure the smoothness of a function by its Lipschitz constant.
	In Appendix~\ref{asec:sobolev}, we extend our results to measure smoothness in terms of  Sobolev norms.
\end{itemize}

Our main results give tight upper- and lower-bounds on the minimum width required for both Lipschitz and Sobolev smooth functions.
The upper- and lower-bounds match up to polynomial factors (equivalently, up to constant factors in the exponent).
The sharpest forms of our bounds involve the number of integer points in certain Euclidean balls; below, we present informal statements of our upper- and lower-bounds for Lipschitz functions with explicit asymptotics given for clarity:

\begin{theorem}[Informal upper-bound for $L$-Lipschitz functions]\label{thm:ub-cube-lipschitz-informal}
	Fix any $\eps, L > 0$ that satisfy $\fracl{L}{\eps} \geq 2$, and let $f: [-1, 1]^d \to \R$ be any $L$-Lipschitz function. For
	\[r =  \exp\paren{O\paren{\min\paren{\frac{L^2}{\eps^2} \log\paren{\frac{d\eps^2}{L^2} + 2}, d \log\paren{\frac{L^2}{\eps^2 d} + 2}}} },\]
with probability $0.9$ (over a draw of $r$ i.i.d.~random ReLU features  $\bgi{1},\dots,\bgi{r}$ from a suitable distribution) there exists a depth-2 RBL ReLU network $h$ with  $\bgi{1},\dots,\bgi{r}$ as the bottom-level features satisfying $\norm[\intvl^d]{f - h} \leq \eps$.
\end{theorem}

\begin{theorem}[Informal lower-bound for $L$-Lipschitz functions]\label{thm:lb-cube-sine-informal}
Fix any $\eps, L > 0$. There exists an $L$-Lipschitz function $f: \intvl^d \to \R$ such that with probability at least $\half$ over a draw of 
	\[r =  \exp\paren{\Omega\paren{\min\paren{\frac{L^2}{\eps^2} \log\paren{\frac{d\eps^2}{L^2} + 2}, d \log\paren{\frac{L^2}{\eps^2 d} + 2}}} }\]
many i.i.d.~random ReLU gates 
 $\bgi{1},\dots,\bgi{r}$, every depth-2 ReLU network $h$ of width $r$ with  $\bgi{1},\dots,\bgi{r}$ as its bottom-layer gates has $\norm[\intvl^d]{f - h} > \eps$.
\end{theorem}

Table~\ref{table:results} summarizes these results, as well as our analogues for functions in Sobolev balls.

\begin{table}[h!]
\centering
\small{
 \begin{tabular}{| l | l | c | l |} 
 \hline
 Bound & Smoothness & Minimum Width & Theorem \\ \hline \hline
 Upper & Lipschitz ${\leq}L$ & $\exp\paren{O\paren{\min\paren{\frac{L^2}{\eps^2} \log\paren{\frac{d\eps^2}{L^2} + 2}, d \log\paren{\frac{L^2}{\eps^2 d} + 2}}} }$ & Thm.~\ref{thm:ub-cube-lipschitz-informal} / \ref{thm:ub-cube-lipschitz} \\ \hline
 Lower & Lipschitz ${\leq}L$ &  $\exp\paren{\Omega\paren{\min\paren{\frac{L^2}{\eps^2} \log\paren{\frac{d\eps^2}{L^2} + 2}, d \log\paren{\frac{L^2}{\eps^2 d} + 2}}} }$ & Thm.~\ref{thm:lb-cube-sine-informal} / \ref{thm:lb-cube-nonexplicit} \\ \hline
 Upper & $H^s$ norm ${\leq}\gamma$ %
       & $\exp\paren{O\paren{\min\paren{d \log\paren{\frac{s \gamma^{2/s}}{d \eps^{2/s}}+2}, \frac{s\gamma^{2/s}}{\eps^{2/s}} \log\paren{\frac{d\eps^{2/s}}{s \gamma^{2/s}}+2}}}}$ & Thm.~\ref{thm:ub-cube-sobolev} \\ \hline
 Lower & $H^s$ norm ${\leq}\gamma$ %
       & $\exp\paren{\Omega\paren{\min\paren{d \log\paren{\frac{\hphantom{s} \gamma^{2/s}}{d \eps^{2/s}}+2}, \frac{\hphantom{s} \gamma^{2/s}}{\eps^{2/s}} \log\paren{\frac{d\eps^{2/s}}{\hphantom{s}\gamma^{2/s}}+2}}}}$ & Thm.~\ref{thm:lb-cube-sobolev} \\ \hline
 \end{tabular}
 }
 \caption{Our upper- and lower-bounds on the minimum width needed for an RBL ReLU network to $\eps$-approximate a function over $\fnmeas{\intvl^d}$ with either bounded Lipschitz constant $L$, or bounded order-$s$ Sobolev norm $\gamma$ (and periodic boundary conditions).  }
\label{table:results}
\end{table}

\noindent {\bf Discussion.}
Our results shed light on a question posed by \cite{ses19} about the approximation power of unconstrained depth-2 networks.
They ask whether there exists a $d$-dimensional 1-Lipschitz function $f$ that can be represented by a depth-3 neural network with $\poly(d)$ neurons but  requires width $\exp(\Omega(d))$ to be approximated by a depth-2 network. As one of their main results, they answer this question in the negative for pointwise approximation when $f$ is a radial function (depending only on $\|x\|_2$) over the unit ball, by showing that any such function can be efficiently approximated by a $\poly(d)$ width depth-2 network. Our results imply that the answer is also negative for $L_2$-approximation of \emph{arbitrary} 1-Lipschitz functions (which need not be radial) over $\intvl^d$; this follows from our upper-bounds for the case that $L = 1$ and $\eps$ is any constant, which establish the existence of approximators that are $\poly(d)$-width, depth-2 RBL networks.
Our results do not answer their question outright, because showing that every 1-Lipschitz function can be approximated with respect to the $L_2$ norm over $\intvl^d$ by a depth-2 network of $\poly(d)$ width does \emph{not} imply that every 1-Lipschitz function is uniformly approximable by such a network.

Our upper-bounds on the width that suffices to approximate Lipschitz functions are also useful for proving learnability hardness results for neural networks with more than two layers.
\cite{myss21} establish this connection between hardness of approximation and hardness of learning by showing that any function that cannot be weakly approximated by a network with three layers cannot be learned by gradient descent applied to a neural network of \emph{any} depth, given certain assumptions about the random weight initialization and bounds on the number of units in the network and number of steps of gradient descent.
Their result hinges on a technical lemma (their Lemma~B.2), which shows that $L$-Lipschitz functions can be approximated by three layer neural networks with bounded width.
By replacing that lemma with our Theorem~\ref{thm:ub-cube-lipschitz}, their result can be strengthened to say that any function not weakly approximable by \emph{two}-layer neural networks is not learnable by gradient descent for networks of any depth that obey their assumptions.

\subsection{Our techniques} \label{sec:techniques}

In this section we give a high-level overview of the ideas that underlie our upper and lower bounds.

\subsubsection{Upper-bounds}

Our width upper-bounds state that for any fixed function of the relevant sort, given a large enough number of independent random ReLU features, with high probability some linear combination of those features approximates the function.
We argue this in three steps. 
(Below, we only discuss the Lipschitzness smoothness measure, but the Sobolev case follows the same basic steps.) 

\begin{enumerate}
  \item The first step shows that for any $L$-Lipschitz function  $f$, there exists a low-degree trigonometric polynomial $P$ that closely approximates $f$.
	We establish the existence of this trigonometric polynomial using the fact that any function in $\fnmeas{\intvl^d}$ can be expressed as a (potentially infinite) linear combination of sinusoidal functions, due to the existence of a Fourier representation for $f$. 
	We use the Lipschitzness of $f$ to show that high-frequency terms have negligibly small coefficients in the representation, which we drop to obtain a low-degree approximation $P$. 

  \item The second step expresses $P$ as an infinite mixture of random ReLU features \citep[\`{a} la][]{barron93,murata1996integral,rubin1998calderon,candes1999harmonic}.
    That is, for some distribution over biases $\bb$ and weights $\bw$ (which depends on $L$, $\eps$, and $d$, but not $f$, and takes values in $\R \times \sph $), $P$ can be written as
	\[P(x) = \EE[\bb,\bw]{h(\bb, \bw) \relu\paren{\innerprod{\bw}{x} - \bb} } \]
  for some function $h(\bb,\bw)$.
  Intuitively, this is possible because each sinusoidal component of $P$ is a ridge function (a function that depends only on a one-dimensional projection of its input).

	\item Finally, using a standard concentration argument, we show that the empirical average of sufficiently many random ReLUs gives a close approximation to $P$ with high probability. %
    It follows that the overall weighted combination of random features closely approximates $f$. 	
	\end{enumerate}

\subsubsection{Lower-bounds}

Our lower-bounds are proved using a dimensionality argument, stemming from the simple observation that linear combinations of $r$ features (functions) can span at most $r$ dimensions in the function space $\fnmeas{\intvl^d}$. 
The key is to give $N \gg r$ candidate functions $\varphi_1, \dots, \varphi_N$ that are orthonormal in $\fnmeas{\intvl^d}$. With such a set of functions in hand, any fixed outcome of a draw of $r$ random features will be such that linear combinations of those $r$ features cannot closely approximate more than a small fraction of the $N$ functions, because no $r$-dimensional subspace can be close to a large fraction of $N$ orthonormal functions. (This kind of dimensionality argument has been used in a number of prior works, including \cite{barron93,ys19,kms20} and elsewhere.)

Specializing to our context, to give a lower-bound on the minimum width of RBL ReLU networks needed to approximate $L$-Lipschitz functions, it suffices to construct a large family of orthonormal $L$-Lipschitz functions. 
We do this with $L$-Lipschitz sinusoidal functions of the form $\sqrt{2} \sin\paren{\pi \innerproduct{K}{x}}$ where $K \in \Z^d$.
The quantity $\|K\|_2$ controls the Lipschitz constant of these functions, and as our analysis shows, the tradeoff between the number of functions in the family (which increases with the allowed range of $\|K\|_2$ and controls our width bound $r$) and the Lipschitz constant $L$ yields a lower-bound that is quite close to our upper-bound for $L$-Lipschitz functions.

The simple dimensionality argument sketched above establishes that some function among the $N$ orthonormal functions is hard to approximate (in fact, that most of them are hard), but it does not yield an \emph{explicit} hard function. By requiring the $N$ orthonormal functions  $\varphi_1, \dots, \varphi_N$ to satisfy a natural symmetry property with respect to the random ReLU features, it is possible to get a lower bound for a single explicit function $\varphi_1$.  Following this approach, we also give a quantitatively slightly weaker lower-bound on the minimum width that random ReLU networks need in order to approximate an explicit function $\varphi_1$.

\subsection{Related work} \label{sec:related-work}

Since the pioneering universal approximation results for (non-RBL) depth-2 networks \citep{cyb89,funahashi1989approximate,hsw89} mentioned in the introduction, many subsequent works have established quantitative bounds on the width that such networks require to approximate certain functions.\footnote{%
  Our discussion here focuses on works that give non-asymptotic bounds.
  \citet[Section 6]{pinkus1999approximation} gives a review of asymptotic rates of approximation by neural networks of width $r$ as $r\to\infty$ (regarding the dimension $d$ as fixed).%
}
RBL networks have also been the subject of considerable study owing to their connection to kernel methods \citep{neal1996priors,rr07,cho2009kernel} and, in particular, the Neural Tangent Kernel (NTK).
\citet{jgh18} argue that training neural networks with gradient descent with small step-sizes results in a learning rule similar to that obtained by a kernel method with the NTK.
When the network weights are randomly initialized, then a finite-width NTK corresponds to a linear combination of random ReLUs.
Both RBL ReLU networks and the finite-width NTK enjoy the same universal approximation property of non-RBL networks~\citep{sgt18,jtx19}, and hence quantitative bounds on the network width required to approximate families of functions are of significant interest.

\paragraph*{Upper-bounds.}
A line of inquiry starting with \cite{barron93} \citep[see also][]{kb18} investigates upper-bounds on the width of (non-RBL) depth-2 networks needed to approximate functions whose smoothness is measured in terms of their Fourier transforms.
Although these results do not deal with RBL networks and hence are incomparable to ours, they do use randomization in the proof.
Specifically, a target function is represented as a mixture of activation functions drawn from a target-specific distribution, and a finite-width depth-2 network approximating the function is obtained by sampling.
Our results use a similar overall approach, but with the crucial difference that in our RBL setting, our distribution of ReLUs does not depend on the target function.

Perhaps the works on RBL networks that are most closely related to our own upper-bounds are those of \citet{apvz}, \citet{ys19}, \citet{bach2017breaking}, and \citet{jtx19}, all of which prove approximation-theoretic results by representing a target function as the expected value of weighted activation functions drawn from some distribution. 
\begin{itemize}
  \item
    Theorem 3.1 of \cite{apvz} shows how neural networks with complex-valued weights and exponential activation functions can approximate polynomials of bounded degree. %
    Their bounds have an exponential dependence on that degree, which translates to an exponential dependence on the Lipschitz constant $L$ even for constant dimension $d$; in contrast, our bounds are exponential in $\min\{d,L^2/\eps^2\}$, which can be much better if $d$ is small.

  \item
    \cite{ys19} study depth-2 RBL ReLU networks (as we do), but like \cite{apvz} focus on approximating polynomials of bounded degree.
    Since they consider a more stringent notion of $L_\infty$-approximation (over the unit ball), their upper-bounds on network width (see their Theorems 3.3 and 3.4) are more pessimistic than ours and depend exponentially on the square of the polynomial degree.

  \item
    Proposition 3 of \cite{bach2017breaking} and Theorem E.1 of \cite{jtx19} imply (or directly give) upper-bounds on the width of depth-2 RBL ReLU networks (or finite-width NTK) to approximate Lipschitz functions.
    Similar to \cite{ys19}, they consider an $L_{\infty}$ notion of approximation, so they obtain upper-bounds that always are exponential in the dimension $d$.

\end{itemize}

\paragraph*{Lower-bounds.}

A number of recent and classical papers give width lower-bounds for arbitrary (non-RBL) depth-2 networks that approximate certain types of multivariate functions.
\cite{maiorov99} gives asymptotically tight upper- and lower-bounds on the error in approximating functions from a Sobolev class achieveable by any two-layer network of a given width.
The asymptotic nature of \citeauthor{maiorov99}'s results (and proof techniques) means that the results do not imply lower-bounds on the network width required to achieve a given error rate $\eps$ unless $\eps$ is sufficiently small, possibly as a function of dimension.
Our results differs from \citeauthor{maiorov99}'s and other related results from the approximation theory literature by elucidating the interplay between the dimension and the error in both upper- and lower-bounds.

More recently, \cite{es15} and \cite{ss16} give $\exp(d)$-type lower-bounds on the width that depth-2 networks require to $L_2$-approximate certain simple functions under certain probability measures on $\R^d$. 
In \cite{es15} the function being approximated is not explicit, and in \cite{ss16} the lower-bound is only for very high-accuracy approximation (to error at most $1/d^4$).
In both works the relevant probability measures are rather involved.
In contrast, our lower bounds hold only for depth-2 RBL networks, but they are for simple explicit functions, for large (constant) values of the approximation parameter, and for $L_2$-approximation with respect to the uniform distribution over $\intvl^d$. 
In other relevant work on depth-2 lower-bounds, \cite{mcpz13} and \cite{dan17-depth} give $\exp(d)$-type (or better) width lower bounds for depth-2 networks approximating certain functions with large Lipschitz constants, but these lower-bounds require a weight bound on the top-level combining gate.
In contrast, our lower bonds for RBL networks have no restrictions on the weights of the top-level gate.

The work of \cite{slchw20}, which analyzes limitations on the approximation abilities of two-layer networks of random ReLU activation functions, is relevant to our lower-bounds.  
Their lower-bounds are independent of the width of the network; they give functions that cannot be approximated by RBL networks of \emph{any} (potentially infinite) width.
However, their lower-bounds are for an extremely strong notion of approximation, namely $L_2$ approximation over all of $\R^d$ (without any weighting by a probability distribution).

Our lower-bound idea of exploiting symmetry to obtain an \emph{explicit} function that is difficult to approximate was inspired by \cite{ys19}.
Our approach for non-explicit lower bounds is quite similar to Theorem 19 of \cite{kms20}, which bounds the dimension of the space of all linear combinations of feature functions; similar to the lower-bound of \cite{kms20} (but unlike \cite{ys19}), our lower-bounds hold regardless of the size of the weights used in the linear combination of the bottom-level random features.

Finally, we remark that while we do not consider networks of depth larger than two, our paper was in large part inspired by results from the literature on depth separation.
\cite{tel16}, \cite{es15}, and \cite{dan17-depth} all prove lower-bounds by constructing highly oscillatory functions and showing that shallow networks must be wide in order to approximate these functions. 
\cite{ses19} prove lower-bounds on 1-Lipschitz functions that are non-oscillatory, such as $x\mapsto\max\{0,-\|x\|+1\}$; however, these bounds only hold in the high-accuracy regime with small $\eps$.
These works motivated us to directly study the relationship between the Lipschitz constant of a target function and the width needed to approximate it.

\section{Preliminaries}\label{sec:prelim}

\subsection{Notations}\label{ssec:notation}

For a positive integer $d \in \Z^+$, let $[d] := \set{1, 2, \dots, d}$. 
The vectors $\vec{0} := (0, \dots, 0) \in \R^d$ and $\vec{1} := (1, \dots, 1) \in \R^d$ are, respectively, the all-zeros and all-ones vectors.
Let $\sph := \setl{x \in \R^d: \norm[2]{x} = 1}$ denote the unit sphere in $\R^d$.
Let $\norm[\lip]{f}$ denote the Lipschitz constant of $f \colon \R^d \to \R$ with respect to the Euclidean metric (i.e., the least $L$ s.t.\ $f$ is $L$-Lipschitz w.r.t.\ $\norm[2]{\cdot}$).

We use the following notations for a multi-index $K \in \N^d$ (where $\N := \setl{z \in \Z : z \geq 0}$).
Let $|K| := \sum_{i=1}^d K_i$, $\norm[2]{K} := (\sum_{i=1}^d K_i^2)^{1/2}$, and $K! := \prod_{i=1}^d (K_i!)$.
Let $x^K := \prod_{i=1}^d x_i^{K_i}$ for $x \in \R^d$.
Lastly, let $\dderiv{K}{f}$ be the order-$|K|$ partial derivative of a function $f(x)$ with respect to $x^K$.

We use bold font to denote random variables and write ``$\bx \sim {\cal D}$'' to indicate that random variable $\bx$ is distributed according to distribution ${\cal D}$.

We use $\innerprodl{\cdot}{\cdot}$ to denote the standard Euclidean inner product in $\R^d$ (and occasionally regard multi-indices $K \in \N^d$ as elements of $\R^d$).
For a probability measure $\mu$ on $\R^d$, $\fnmeas{\mu}$ denotes the space of square-integrable functions with inner product denoted by $\ipmeasl{f}{g}{\mu} := \EEl[\bx \sim\mu]{f(\bx) g(\bx)} = \int_{\R^d} f(x) g(x) \mu(\dif x)$.
Many of our results concern the uniform probability measure on $\intvl^d$.
In these cases, we use the notations $\fnmeas{\intvl^d}$ and $\ipmeasl{\cdot}{\cdot}{\intvl^d}$, and fix a particular orthonormal basis $\mathcal{T} = \setl{ T_K : K \in \Z^d }$ for $\fnmeas{\intvl^d}$ based on trigonometric polynomials.
See Appendix~\ref{sec:trig-basis} for details.
We also consider certain finite-dimensional subspaces of $\fnmeas{\intvl^d}$ which are spanned by a set of functions indexed by $\mathcal{K}_{k,d} := \setl{K \in \Z^d: \norm[2]{K} \leq k}$.
The dimensions $Q_{k,d} := |\mathcal{K}_{k,d}|$ of these subspaces are upper- and lower-bounded as follows (proof also given in Appendix~\ref{sec:trig-basis}).
\begin{fact}\label{fact:Qkd-ub}
	For all $d \in \Z^+$ and $k \geq 1$,  $Q_{k, d} = \exp\paren{\Theta\paren{\min\paren{d \log\paren{\frac{k^2}{d}+2}, k^2 \log\paren{\frac{d}{k^2}+2}}}}$.
\end{fact}

\subsection{Random bottom layer neural network approximation}\label{ssec-rbl}

Throughout the paper, we treat a depth-2 random bottom layer (RBL) ReLU network as a random features model.
The upper-bounds in this paper demonstrate the representational powers of linear combinations of these random features, while the lower-bounds demonstrate their limitations. 

We define a family of distributions over the parameters of random ReLU activations. 
Note that our lower-bounds in Theorems~\ref{thm:lb-cube-nonexplicit}, \ref{thm:lb-cube-sine}, \ref{thm:lb-cube-sobolev}, and~\ref{thm:lb-cube-sobolev-explicit} hold for \emph{all} such distributions $\dprod$, while our upper-bounds in Theorems \ref{thm:ub-cube-lipschitz} and \ref{thm:ub-cube-sobolev} hold for some fixed $\dprod$, which depends on an upper bound on the Lipschitz norm of the target function but not on the target function itself.

\begin{definition}[Symmetric ReLU parameter distributions]\label{def:sym-relu-param-dist}
A product distribution $\dprod := \dbias \times \dweight$ over $\R \times \sph$ is a \emph{symmetric ReLU parameter distribution} if the coordinates of $\dweight$ are invariant to permutation. That is, $\dweight = \pi \circ \dweight$ for any permutation $\pi$ of $[d]$. \end{definition}

Given a distribution over random ReLU parameters, we now introduce the full random ReLU features model.
We define a notion of approximation and formalize the \emph{minimum width} of the network (or the minimum number of random features to combine) needed to obtain a sufficiently accurate approximation with high probability.

\begin{definition}[Minimum-width RBL ReLU network approximation]\label{def:minwidth}
Consider a symmetric ReLU parameter distribution $\dprod$, a measure $\mu$ over $\R^d$, and a network width $r \in \Z^+$.
For all $i \in [r]$, we draw each \emph{random network feature} $\bgi{i} \in \fnmeas{\mu}$ independently by drawing $(\bbi{i}, \bwi{i})$ from $\mathcal{D}$ and letting $\bgi{i}(x) := \relu(\innerprodl{\bwi{i}}{x} - \bbi{i})$.

Given $\eps, \delta > 0$ and a function $f: \R^d \to \R$ with bounded $\norm[\mu]{f}$, we define $\minwidth{f}{\eps}{\delta}{\mu}{\mathcal{D}}$ to be the smallest $r \in \Z^+$ such that the following holds: With probability at least $1 - \delta$ over $\bgi{1}, \dots, \bgi{r}$, 
	\[\inf_{g \in \Span{\bgi{1}, \dots, \bgi{r}}}\norm[\mu]{f - g} \leq \eps.\]
\end{definition}

\section{Upper-bounds  for Lipschitz functions in $\fnmeas{\intvl^d}$}\label{sec:ub-cube-lipschitz}
Our upper-bounds on the minimum width RBL ReLU network that approximates a Lipschitz function are dominated by the quantity $Q_{k, d}$, which represents the number of integer points contained in a $d$-dimensional ball of radius $k$ (see Section~\ref{ssec:notation}).

\begin{theorem}[Formal version of Theorem \ref{thm:ub-cube-lipschitz-informal}: Upper-bound for $L$-Lipschitz functions]\label{thm:ub-cube-lipschitz}
	Fix some $\delta \in (0, \half]$ and $\eps, L > 0$ with $\frac{L}{\eps} \geq 2$.
	Then, there exists some symmetric ReLU parameter distribution $\mathcal{D}$ such that for any $f \in \fnmeas{\intvl^d}$ with $\norm[\lip]{f} \leq L$ and $\abs{\EE[\bx]{f(\bx)} } \leq L$,  
	\[\minwidth{f}{\eps}{\delta}{\intvl^d}{\mathcal{D}} \leq O\paren{\frac{ L^6 d^2 }{\eps^6} \ln \paren{\frac{1}{\delta}}Q_{2L / \eps,d}^2}.\]
\end{theorem}
Applying the asymptotics of $Q_{k,d}$ from Fact~\ref{fact:Qkd-ub} reveals that the minimum width can also be bounded by the term in Theorem~\ref{thm:ub-cube-lipschitz-informal}.
That expression shows that the minimum width is polynomial in $\frac{L}{\eps}$ when $d$ is a fixed constant, and polynomial in $d$ when $\frac{L}{\eps}$ is a fixed constant. 

To prove Theorem~\ref{thm:ub-cube-lipschitz}, we break the process of approximating a Lipschitz function $f$ with an RBL ReLU network into two steps.
We first approximate $f$ with a bounded-degree trigonometric polynomial $P$ in Lemma~\ref{lemma:trig-lipschitz-approx} and then approximate $P$ with an RBL ReLU network in Lemma~\ref{lemma:ub-cube-trig-bounded-degree}. 
We state the lemmas and discuss their proofs in Sections~\ref{ssec:trig-lipschitz} and~\ref{ssec:ub-cube-trig-bounded-degree} respectively.
Section~\ref{ssec:ub-cube-lipschitz-proof} gives a formal proof of Theorem~\ref{thm:ub-cube-lipschitz}.

In Appendix~\ref{asec:ub-cube-sobolev}, we present and prove Theorem~\ref{thm:ub-cube-sobolev}, a parallel result to Theorem~\ref{thm:ub-cube-lipschitz} that instead considers the approximation of some function $f$ that has a bounded Sobolev norm and which (along with its derivatives) satisfies periodic boundary conditions.
The proof of  Theorem~\ref{thm:ub-cube-sobolev} only differs from that of Theorem~\ref{thm:ub-cube-lipschitz} by obtaining a trigonometric polynomial approximation for $f$ from Lemma~\ref{lemma:trig-sobolev-approx} (stated and proved in Appendix~\ref{asec:ub-cube-sobolev}) rather than Lemma~\ref{lemma:trig-lipschitz-approx}.

\subsection{Approximating Lipschitz functions with bounded-degree trigonometric polynomials}\label{ssec:trig-lipschitz}
\begin{lemma}\label{lemma:trig-lipschitz-approx}
	Fix some $L, \eps > 0$ with $\frac{L}{\eps} \geq 1$ and consider any function $f \in L^2([-1, 1]^d)$ with $\norm[\lip]{f} \leq L$ and $\abs{\EE[\bx]{f(\bx)}} \leq L$. 
	Then, taking $k = {\frac{L}{\eps}}$, there exists a bounded-degree trigonometric polynomial \[P(x) = \sum_{K \in \mathcal{K}_{k,d}} \beta_K T_{K}\paren{\frac{x}{2}} \]
	such that 
	$\norm[\intvl^d]{f - P} \leq \eps.$
	Moreover, $\abs{\beta_K} \leq L$ for all $K$. \end{lemma}
	We formally prove this lemma (which we restate as Lemma~\ref{lemma:trig-lipschitz-approx-restate}) in Appendix~\ref{ssec:trig-lipschitz-approx}.
Here we highlight a central part of the argument (used in the full proof) by stating and proving a special case of the lemma which additionally requires that $f$ satisfy periodic boundary conditions.
\begin{lemma}[Approximating Lipschitz functions with periodic boundary conditions]\label{lemma:trig-lipschitz-approx-periodic} \ %
	Fix some $L, \eps > 0$ with $\frac{L}{\eps} \geq 2$. 
	Consider any function $f \in L^2([-1, 1]^d)$ such that $f$ satisfies periodic boundary conditions, $\norm[\lip]{f} \leq L$, and $\abs{\EE[\bx]{f(\bx)}} \leq \frac{L}{2}$. 
	Then, taking $k = {\frac {L}{2 \eps}}$, there exists a bounded-degree trigonometric polynomial \[P(x) = \sum_{K \in \mathcal{K}_{k,d}} \beta_K T_{K}\paren{x} \]
	such that 
	$\norm[\intvl^d]{f - P} \leq \eps.$
	Moreover, $\abs{\beta_K} \leq \frac{L}{2}$ for all $K$. \end{lemma}
	To prove Lemma~\ref{lemma:trig-lipschitz-approx-periodic}, we consider the representation of $f$ as an infinite linear combination of trigonometric basis elements from $\mathcal{T}$.
	We show that $f$ can only be $L$-Lipschitz if all high-degree terms of this representation have vanishingly small coefficients.
	This requires the term-by-term differentiation of the trigonometric representation of $f$, which is possible due to its periodic boundary conditions (see Lemma~\ref{lemma:trig-expansion-deriv} in Appendix~\ref{sec:trig-basis}).

\begin{proofnoqed}
  By appealing to a standard approximation argument~\citep[e.g.,][Proposition 8.17]{folland1999real}, we may assume that $f$ is differentiable. %
	Because $\mathcal{T}$ is an orthonormal basis over $\fnmeas{[-1, 1]^d}$, we can express $f$ as 
	\[{f}(x) = \sum_{K \in \Z^d} {\alpha}_K T_K(x).\]
	The condition $\norm[\lip]{f} \leq L$ implies that $\norm[2]{\nabla f(x)} \leq L$  for all $x \in \intvl^d$.
	Because $f$ has periodic boundary conditions, $f$ is differentiable, and $\pderivl{{f}(x)}{x_i} \in \fnmeas{\intvl^d}$ for all $i$, Lemma~\ref{lemma:trig-expansion-deriv} can be applied to relate $L$ to the coefficients $(\alpha_K)_{K \in \Z^d}$:
	\begin{align*}
		L^2
		&\geq \EE[\bx \sim \intvl^d]{\norm[2]{\nabla {f}(\bx)}^2} = \sum_{i=1}^d \EE[\bx]{\paren{\pderiv{{f}(\bx)}{\bx_i}}^2} = \sum_{i=1}^d \EE[\bx]{\paren{\sum_{K \in \Z^d} {\alpha}_K \pderiv{T_K(\bx)}{\bx_i}}^2}  \stepcounter{equation}\tag{\theequation}\label{line:nnn} \\
		&= \sum_{i=1}^d \sum_{K \in \Z^d} {\alpha}_K^2 \norm[\intvl^d]{ \pderiv{T_K}{x_i}}^2 + 2  \sum_{i=1}^d \sum_{K \in \Z^d} \sum_{K' \neq K}{\alpha}_K {\alpha}_{K'} \ipmeas{ \pderiv{T_K}{x_i}}{ \pderiv{T_{K'}}{x_i}}{\intvl^d} \\
		&= \sum_{i=1}^d  \sum_{K \in \Z^d} {\alpha}_K^2 \pi^2 K_i^2 \stepcounter{equation}\tag{\theequation}\label{line:f} = \pi^2 \sum_{K}{\alpha}_K^2 \norm[2]{K}^2.
	\end{align*}
	
	Equations \eqref{line:nnn} and \eqref{line:f} follow from Lemma~\ref{lemma:trig-expansion-deriv} and Fact~\ref{fact:trig-pderiv} respectively. 
	An immediate consequence of the above inequality is that $\abs{{\alpha}_K} \leq \fracl{L}{\pi} \leq \fracl{L}{2}$ as long as $K \neq \vec{0}$. 
	Because $\abs{\EE[\bx]{{f}(\bx)}} \leq \fracl{L}{2}$, $\absl{{\alpha}_{\vec{0}}} \leq \fracl{L}{2}$ as well.
	We define the trigonometric polynomial ${P} = \sum_{K \in \mathcal{K}_{k,d}} {\beta}_K T_K$ by letting ${\beta}_K := {\alpha}_K$ for all $K$ with $\norm[2]{K} \leq k$. Parseval's identity (Fact \ref{fact:parseval-plancherel}) and the inequality ending on line \eqref{line:f} guarantee that
	\begin{align*}
		\norm[\intvl^d]{{f} - {P}}^2 
		&= \sum_{K \in \Z^d \setminus \mathcal{K}_{k,d}} {\alpha}_K^2 
		\leq \sum_{K \in \Z^d \setminus \mathcal{K}_{k,d}} {\alpha}_K^2 \cdot \frac{\norm[2]{K}^2}{k^2} 
		\leq  \frac{1}{k^2}\sum_{K \in \Z^d} {\alpha}_K^2 \norm[2]{K}^2 \\ 
		&\leq \frac{L^2}{\pi^2 k^2} 
		\leq \frac{L^2}{2^2 k^2} 
		= \eps^2.
    \qedhere
	\end{align*}
\end{proofnoqed}

The proof of Lemma~\ref{lemma:trig-lipschitz-approx} is a reduction to Lemma~\ref{lemma:trig-lipschitz-approx-periodic}.
	Instead of approximating $f$ with a low-degree trigonometric polynomial, we approximate $\tilde{f}$, a scaled, shifted, and reflected version of $f$ that has periodic boundary conditions and thus can be differentiated term-by-term. 
	The bulk of the proof involves transforming $f$ into $\tilde{f}$ and transforming $\tilde{P}$ (the trigonometric polynomial obtained by applying Lemma~\ref{lemma:trig-lipschitz-approx-periodic} to $\tilde{f}$) back into $P$.
	This scaling and reflection argument is why we approximate $f$ with combinations of trigonometric polynomials of the form $T_K(\fracl{x}{2})$, rather than $T_K(x)$.

\subsection{Approximating bounded-degree trigonometric polynomials with RBL ReLU nets}\label{ssec:ub-cube-trig-bounded-degree}
\begin{lemma}\label{lemma:ub-cube-trig-bounded-degree}
	Fix some $\delta \in (0, \halfl]$, $\eps > 0$, $\rho \in (0, 1]$, $k \geq 1$, and $d \in \mathbb{Z}^+$. 
	Then, there exists some symmetric ReLU parameter distribution $\mathcal{D}_k$ such that 
	for any trigonometric polynomial \[P(x) = \sum_{K \in \mathcal{K}_{k,d}}\beta_K T_K(\rho x)\] with $\abs{\beta_K} \leq \betam$ for all $K \in \mathcal{K}_{k,d}$, 
	\[\minwidth{P}{\eps}{\delta}{\intvl^d}{\mathcal{D}_k} \leq O\paren{\frac{ \betam^2 d^2 k^4 }{\eps^2} Q_{k,d}^2  \ln \paren{\frac{1}{\delta}}}.\]

\end{lemma}

	We prove this lemma in Appendix~\ref{ssec:ub-cube-trig-bounded-degree-app} as Lemma~\ref{lemma:ub-cube-trig-bounded-degree-restate}. 
	We take advantage of the fact that every low-degree trigonometric polynomial can be expressed as a linear combination of ridge functions.
	As shown in Lemma \ref{lemma:relu-univariate-representation}, each of those ridge functions can in turn be represented as an infinite mixture of ReLUs.
  We then represent the entire trigonometric polynomial as an expectation over weighted random ReLU features with parameters drawn from a symmetric ReLU parameter distribution $\mathcal{D}_k$ (Definition \ref{def:Dk}).
	By bounding the maximum norm of every random ReLU drawn from $\mathcal{D}_k$, a concentration bound (Lemma \ref{lemma:rr08}) can show that this expectation can be closely approximated with a sufficiently large finite linear combination of randomly sampled ReLUs. 
		
\subsection{Proof of Theorem \ref{thm:ub-cube-lipschitz}}\label{ssec:ub-cube-lipschitz-proof}
	Consider any $f \in \fnmeas{\intvl^d}$ with $\norm[\lip]{f} \leq L$ and $\abs{\EE[\bx]{f(\bx)}} \leq L$. 
	By Lemma \ref{lemma:trig-lipschitz-approx}, there exists a bounded-degree trigonometric polynomial
	$P(x) = \sum_{K \in \mathcal{K}_{k,d}} \beta_K T_K\paren{\fracl{x}{2}}$
	with $k := \fracl{2L}{\eps}$ and $\abs{\beta_K} \leq L$ for all $K \in \mathcal{K}_{k,d}$, such that
	$\norm[\intvl^d]{f - P} \leq \fracl{\eps}{2}.$
	By applying Lemma \ref{lemma:ub-cube-trig-bounded-degree} to $P$ with $\rho = \fracl{1}{2}$,	\begin{align*}
	\minwidth{P}{\eps/2}{\delta}{\intvl^d}{\mathcal{D}_k} 
	\leq O\paren{\frac{ \betam^2 d^2 k^4 }{\eps^2} Q_{k,d}^2 \ln \paren{\frac{1}{\delta}} }
	&\leq  O\paren{\frac{ d^2 L^6}{\eps^6}Q_{2L/ \eps,d}^2  \ln \paren{\frac{1}{\delta}}}.
	\end{align*}
	Thus (see Definition \ref{def:minwidth}) there exists an RBL ReLU network $g$ of width $\minwidth{P}{\eps/2}{\delta}{\intvl^d}{\mathcal{D}_k}$  such that 
	$\norm[\intvl^d]{P - g} \leq \fracl{\eps}{2}.$
	By the triangle inequality,
	$\norm[\intvl^d]{f - g} \leq \eps.$
	We conclude that 
	\begin{align*}
		\minwidth{f}{\eps}{\delta}{\intvl^d}{\mathcal{D}_k} 
		&= O\paren{\frac{ d^2 L^6}{\eps^6} Q_{2L/ \eps,d}^2 \ln \paren{\frac{1}{\delta}} } .
    \qedhere
	\end{align*}

\section{Lower-bounds for Lipschitz functions in $\fnmeas{\intvl^d}$} \label{sec:lb}

We give lower-bounds on the minimum width needed to $\eps$-approximate $L$-Lipschitz functions using depth-2 RBL ReLU networks.
Below we present a formal statement of Theorem \ref{thm:lb-cube-sine-informal}, which shows that a particular family of ``simple'' functions must contain some hard-to-approximate function.
Like the upper-bounds in Section~\ref{sec:ub-cube-lipschitz}, the minimum width is polynomial (in fact linear) in the quantity $Q_{k,d}$, where $k = \Theta\parenl{\fracl{L}{\eps}}$.

\begin{theorem}[Formal version of Theorem \ref{thm:lb-cube-sine-informal}: Lower-bound for $L$-Lipschitz functions]\label{thm:lb-cube-nonexplicit}
	Fix any $\eps, L > 0$ and fix any symmetric ReLU parameter distribution $\dprod$. Then, there exists some multi-index $K \in \N^d$ with $\norm[2]{K} \leq \fracl{L}{18\eps}$ such that the function
	$f(x) := 4 \eps T_K$ (recall that $T_K \in \mathcal{T}$) satisfies
	$\norm[\lip]{f} \leq L$ and
	\[\minwidth{f}{\eps}{\half}{\intvl^d}{\mathcal{D}} \geq \frac{1}{4} Q_{L/18\eps,d} . \]
\end{theorem}
The informal version, Theorem~\ref{thm:lb-cube-sine-informal}, follows by 
applying Fact~\ref{fact:Qkd-ub} to lower-bound $Q_{k,d}$.
We note that the function $f$ used in the lower-bound aligns nicely with the approximation techniques from Section~\ref{sec:ub-cube-lipschitz} because $f$ is \emph{(i)} a ridge function and \emph{(ii)} a scalar multiple of a sinusoidal function from the trigonometric basis $\mathcal{T}$. 

We prove Theorem~\ref{thm:lb-cube-nonexplicit} in stages by proving a sequence of claims which are successively more closely tailored to our RBL ReLU model.
\begin{enumerate}
	\item 
	In Appendix~\ref{ssec:general-lb-app} we state and prove Theorem~\ref{thm:randict-simplified}, which gives a general result about the limitations of linear combinations of $r$ random features. This theorem states that a large fraction of any set of $N$ orthonormal functions must be inapproximable by linear combinations of $r$ random features when $N \gg r$.
	We state a simplified version of the theorem below:
	\begin{theorem}[Simplification of Theorem~\ref{thm:randict}]\label{thm:randict-simplified}
  Let $\Phi = \set{\varphi_1,\dotsc,\varphi_N} \subset \fnmeas{\mu}$ be a family of $N$ functions such that $\ipmeas{\varphi_i}{\varphi_{i'}}{\mu} = \indicator{i =i'}$.
  Let $\bgi{1},\dotsc,\bgi{r}$ be i.i.d.~copies of an $\fnmeas{\mu}$-valued random variable.
  Then, there exists some $\varphi_i \in \Phi$ such that 
  \begin{equation*}
    \EE[\bgi{1},\dotsc,\bgi{r}]{ \inf_{g \in \spangl} \norm[\mu]{g - \varphi_i}^2 }
    \geq 1 - \frac{r }{N} .
 \end{equation*}
\end{theorem}
	The proof hinges on an intuitive linear algebraic fact generalized to function spaces: $N$ orthogonal vectors cannot all be close to the span of $r$ vectors when $N \gg r$. 
	It does so by applying the Hilbert Projection Theorem (Fact~\ref{fact:hilbert-proj}).
	The full generality of Theorem~\ref{thm:randict} also includes function families $\Phi$ that are ``nearly orthonormal'' rather than strictly orthonormal (this generalization is useful for extending our results to Gaussian space, as discussed in Appendix~\ref{asec-gaussian}).
	It also proves the inapproximability of some explicit function $\varphi_1$ when the family $\Phi$ satisfies a suitable notion of symmetry relative to $\bgi{1},\dotsc,\bgi{r}$.

	\item 
	Lemma~\ref{lemma:randict-relu} of Appendix~\ref{ssec:lb-relu-app} adapts Theorem~\ref{thm:randict} to our random ReLU features by giving a lower-bound on the minimum width RBL network needed to $\eps$-approximate some function for any $\eps > 0$.
	Below is a simplified version of the lemma that is restricted to orthonormal function families, considers only the uniform measure over $\intvl^d$, and omits the special ``symmetric case'' for $\Phi$.
	
	\begin{lemma}[Simplification of Lemma~\ref{lemma:randict-relu}]\label{lemma:randict-relu-simplified}
Let $\mathcal{D}$ be a symmetric ReLU parameter distribution. Fix any $\Phi = \set{\varphi_1,\dotsc,\varphi_N} \subset \fnmeas{\intvl^d}$ such that $\ipmeasl{\varphi_i}{\varphi_{i'}}{\intvl^d} = \indicator{i = i'}$.
Then, for any $\eps > 0$, there exists some $\varphi_i \in \Phi$ such that
$\minwidth{4 \eps \varphi_{i}}{\eps}{\halfl}{\intvl^d}{\mathcal{D}} \geq \fracl{N}{4} .$
\end{lemma}
	The proof combines a scaling argument with the definition of $\mathrm{MinWidth}$ to provide lower-bounds for any choice of the error parameter $\eps$. 

	\item We conclude the proof of Theorem~\ref{thm:lb-cube-nonexplicit} in Appendix~\ref{ssec:lb-relu-cube-tight-app}.
	Lemma~\ref{lemma:lb-cube-nonexplicit} shows the existence of a low-degree element of the sinusoidal basis $\mathcal{T}$ that cannot be approximated over $\intvl^d$ by an RBL ReLU network of small width. 
	It does so by defining the orthonormal family of functions to be $\Phi := \setl{T_K \in \mathcal{T}: K \in \mathcal{K}_{k,d}}$ and invoking Lemma~\ref{lemma:randict-relu}.
	The proof of Theorem~\ref{thm:lb-cube-nonexplicit} only requires applying Lemma~\ref{lemma:lb-cube-nonexplicit}  for some $k = \Theta\parenl{\fracl{L}{\eps}}$ and showing that all $T_K \in \Phi$ have $\norm[\lip]{T_K} \leq L$.
	
	Lemma~\ref{lemma:lb-cube-nonexplicit} also yields an immediate proof of Theorem~\ref{thm:lb-cube-sobolev}, the Sobolev analogue of Theorem~\ref{thm:lb-cube-nonexplicit},  in Appendix~\ref{asec:lb-cube-sobolev}. Theorem~\ref{thm:lb-cube-sobolev} uses the same function family $\Phi$, but must bound the Sobolev norm of all functions in $\Phi$ rather than the Lipschitz constant.
\end{enumerate}

The lower-bound established in Theorem~\ref{thm:lb-cube-nonexplicit} is non-explicit; it guarantees the existence of some inapproximable function in $\mathcal{T}$, but does not by itself let us deduce the specific identity of a hard function.  
Since it is desirable to have a lower-bound for a fully explicit function, we also give a variant that achieves this goal at only a small cost in the resulting quantitative lower-bound:

\begin{theorem}[Explicit lower-bound for an $L$-Lipschitz function]\label{thm:lb-cube-sine}
	For some $\eps, L > 0$, let $\ell := {\min\parenl{\ceill{\fracl{d}{2}}, \floorl{\fracl{L^2}{32 \pi^2 \eps^2}}}}$. Fix any symmetric ReLU parameter distribution $\dprod$. Then the function $f(x) := 4 \sqrt{2} \eps \sin\parenl{\pi \sum_{i=1}^{\ell} x_i}$
	satisfies $\norm[\lip]{f} \leq L$ and
	\[\minwidth{f}{\eps}{\half}{\intvl^d}{\mathcal{D}} \geq \frac{1}{4} {d \choose \ell}  \geq \exp\paren{\Omega\paren{\min\paren{\frac{L^2}{\eps^2}\log\paren{\frac{d\eps^2}{L^2}+2}, d}}}.\]
\end{theorem}
Comparing the quantitative lower-bounds of Theorem \ref{thm:lb-cube-nonexplicit} and Theorem \ref{thm:lb-cube-sine}, we see that the latter is weaker only by a logarithmic factor in the exponent.

We prove the explicit lower-bound Theorem \ref{thm:lb-cube-sine} in Appendix~\ref{ssec:lb-relu-cube-explicit-app}.
The only difference between the proofs of Theorems~\ref{thm:lb-cube-nonexplicit} and~\ref{thm:lb-cube-sine} is in the last step.
Theorem~\ref{thm:lb-cube-sine} relies on Lemma~\ref{lemma:lb-cube-sine}, an analogue of Lemma \ref{lemma:lb-cube-nonexplicit}, which invokes Lemma \ref{lemma:randict-relu} with a different family $\Phi$ of trigonometric polynomials that are symmetric up to a permutation of variables.
That is, for every $T_K, T_{K'} \in \Phi$, there exists some permutation $\pi$ over $[d]$ such that $T_K = T_{K'} \circ \pi$. 
(Roughly speaking, the larger family of orthonormal functions used in the proof of Theorem~\ref{thm:lb-cube-nonexplicit} consists of functions of the form $\sin\paren{\pi \innerprod{K}{x}}$ where $K \in \N^d$ is only constrained by having $\|K\|$ satisfy some bound, whereas the smaller family of orthonormal functions used in the proof of Theorem~\ref{lemma:lb-cube-sine} consists of functions of the form $\sin\paren{\pi \innerprod{K}{x}}$ where $K $ is restricted to be a 0/1 vector of some specific Hamming weight. 
The latter family is easily seen to satisfy symmetry with respect to any permutation $\pi$ of the $d$ coordinates, whereas the former family does not satisfy such a symmetry condition.) 
This symmetry condition makes it easy to argue that all functions in the symmetric family $ \Phi$ are ``equally hard,'' from which a lower bound follows straightforwardly.

Finally, we mention that Lemma~\ref{lemma:lb-cube-sine} also supports a proof of the inapproximability of an explicit function with bounded Sobolev norm; this is established in Theorem~\ref{thm:lb-cube-sobolev-explicit} of Appendix~\ref{asec:lb-cube-sobolev}.

\acks{In this work, D. Hsu is supported by NSF grants CCF-1740833, IIS-1563785, and a Sloan Research Fellowship.
C. Sanford gratefully acknowledges CCF-1563155 and is partially supported by Google Faculty Research Award.
R.A. Servedio is supported by NSF grants CCF-1814873, IIS-1838154, CCF-1563155, and by the Simons Collaboration on Algorithms and Geometry.
E.V. Vlatakis-Gkaragkounis is grateful to be supported by NSF grants CCF-1703925, CCF-1763970, CCF-1814873, CCF-1563155, and by the Simons Collaboration on Algorithms and Geometry and by the Onassis Foundation under Scholarship ID: F ZN 010-1/2017-2018.
Finally, the authors would like to thank Shivam Nadimpalli for the helpful discussions about the necessary conditions of Lemma~\ref{lemma:trig-expansion-deriv}. 
This material is based upon work supported by the National Science Foundation under grant numbers listed above. Any opinions, findings and conclusions or recommendations expressed in this material are those of the authors and do not necessarily reflect the views of the National Science Foundation (NSF).
}
\bibliography{ourbibliography}
\appendix
\section{Key facts about trigonometric polynomial basis}\label{sec:trig-basis}

In this appendix, we supplement Section~\ref{ssec:notation} by introducing the family of trigonometric polynomials that we use in our proofs and by proving properties related to their orthonormality. 
We recall the definition of an orthonormal basis for the space  $L_2(\mu)$:
\begin{definition}[Orthonormal basis]
	A countable set $\mathcal{G} \subset \fnmeas{\mu}$ is an \emph{orthonormal basis} for $\fnmeas{\mu}$ if  $\ipmeas{g}{\tilde{g}}{\mu} = \indicator{g = \tilde{g}}$ for all $g, \tilde{g}\in \mathcal{G}$ and $\Span{\mathcal{G}}=\fnmeas{\mu}$.
\end{definition} 
We frequently apply the following standard facts about orthonormal bases:
\begin{fact}[Facts about orthonormal bases]\label{fact:parseval-plancherel}
	For some measure $\mu$, let $\mathcal{G}$ be an orthonormal basis for $\fnmeas{\mu}$. For any $f, \tilde{f} \in \fnmeas{\mu}$ we have that $f = \sum_{g \in \mathcal{G}} \alpha_g g$ and $\tilde{f} = \sum_{g \in \mathcal{G}} \beta_g g$ for some real $(\alpha_g)_{g \in \mathcal{G}}$ and $(\beta_g)_{g \in \mathcal{G}}$, and moreover
	\begin{itemize}
		\item $\alpha_g = \ipmeasl{f}{g}{\mu}$;
		\item $\norm[\mu]{f}^2 = \sum_{g \in \mathcal{G}} \alpha_g^2$ (Parseval); and
		\item $\ipmeasl{f}{\tilde{f}}{\mu} = \sum_{g \in \mathcal{G}} \alpha_g \beta_g$ (Plancherel).
	\end{itemize}
\end{fact}

We define the basis of trigonometric polynomials $\mathcal{T}$ as
\[\mathcal{T} := \set{T_K: K \in \Z^d},\]
where
\begin{equation}
T_K(x) := \begin{cases}
	1 & K = \vec{0} \\
	\sqrt{2}\sin\paren{\pi \innerprod{K}{x}} & K \in \sinset \\
	\sqrt{2}\cos\paren{\pi \innerprod{K}{x}} & K \in \cosset ,
\end{cases}
\label{eq:Tk} 
\end{equation}
and $\sinset$ and $\cosset$ form a partition of $\Z^d \setminus \setl{\vec{0}}$\footnote{
Note that this partition of $\Z^d - \setl{\vec{0}}$ is an arbitrary one. 
The only property this partition is designed to satisfy is that if $K$ corresponds to $\sin(\pi \innerprodl{K}{ x})$, then $-K$ must correspond to $\cos(-\pi \innerprodl{K}{x})$ (and vice versa). 
}
and are defined as
\begin{align*}
	\sinset &:= \set{K \in \Z^d \setminus \setl{\vec{0}}: K_i > 0, \ \text{where $i = \min \set{j \in [d]: x_j \neq 0}$}}, \\
	\cosset &:= \set{K \in \Z^d \setminus \setl{\vec{0}}: K_i < 0, \ \text{where $i = \min \set{j \in [d]: x_j \neq 0}$}}.
\end{align*}
The set
$\mathcal{T}$ is a useful family of functions for both our upper- and our lower-bounds on the minimum width RBL ReLU network needed to approximate Lipschitz functions. 
The fact that $\mathcal{T}$ is an orthonormal basis for $\fnmeas{\intvl^d}$ (Fact \ref{fact:trig-orthonormal-basis-cube}) permits us to express other functions in $\fnmeas{\intvl^d}$ as a linear combination of the elements of $\mathcal{T}$.
As we show in Fact~\ref{fact:trig-pderiv}, those orthogonality properties of the elements of $\mathcal{T}$ are maintained even after taking partial derivatives.
In addition, every function in $\mathcal{T}$ is a ridge function (that is, $T_K(x) = \phi_K(\innerprod{K}{x})$ for some $\phi_K: \R \to \R$), which, as we will see later, means (very usefully for us) that $T_K$ is easily approximated by linear combinations of shifted ReLUs.
Finally, the Lipschitz constant of all functions in $\mathcal{T}$ is bounded: $\norm[\lip]{T_K} \leq \sqrt{2} \pi \norm[2]{K}$. 

To prove that $\mathcal{T}$ is orthogonal, we rely on the following fact from integral calculus.
\begin{fact}[Integrals of multivariate sinusoids]\label{fact:multi-integral-cosine}
	For each $K \in \mathbb{Z}^d$, 
  \begin{align*}
    \int_{[-1, 1]^d} \cos\paren{\pi \innerprod{K}{x}} \dif x
    & = 2^d \cdot \indicatorl{K = \vec{0}}\quad \& \quad \int_{[-1, 1]^d} \sin\paren{\pi \innerprod{K}{x}} \dif x
     = 0 .
  \end{align*}
\end{fact}
\begin{proof}
 We use a simple inductive argument on $d$ to evaluate the first integral.
 The base case $d=1$ is straightforward, so assume $d>1$ and
 define $x_{-1} = (x_2, \dots, x_d) \in \R^{d-1}$ for any $x \in \R^d$.
 Assume inductively that
	\[\displaystyle\int_{[-1, 1]^{d-1}} \cos\paren{\pi \innerprod{K_{-1}}{x_{-1}}} \dif x_{-1} = 2^{d-1} \indicatorl{K_{-1} = \vec{0}}.\]
	By the cosine addition formula, we have that:
	\begin{align*}
    \lefteqn{ \int_{[-1, 1]^d} \cos\paren{\pi{\innerprod{K}{x}}} \dif x } \\
  & = \int_{[-1, 1]^d} \bracket{\cos\paren{\pi K_1 x_1}\cos\paren{\pi{\innerprod{K_{-1}}{x_{-1}}}} - \sin\paren{\pi K_1 x_1}\sin\paren{\pi{\innerprod{K_{-1}}{x_{-1}}}}} \dif x \\
		& =  \bracket{\int_{-1}^1\cos\paren{\pi K_1 x_1} \dif x_1}\bracket{\int_{[-1,1]^{d-1}}\cos\paren{\pi \innerprod{K_{-1}}{x_{-1}}}\dif x_{-1}} \\
    & \qquad -\bracket{\int_{-1}^1\sin\paren{\pi K_1 x_1} \dif x_1} \bracket{\int_{[-1,1]^{d-1}}\sin\paren{\pi \innerprod{K_{-1}}{x_{-1}}}\dif x_{-1}} \\
		& = 2 \cdot \indicatorl{K_1 = 0} \bracket{\int_{[-1,1]^{d-1}}\cos\paren{\pi \innerprod{K_{-1}}{x_{-1}}}\dif x_{-1}} %
		= 2^d \cdot \indicatorl{K = \vec{0}}.
	\end{align*}
	The second claim follows by a nearly identical inductive argument, which we omit.
	\end{proof}	

\begin{fact}\label{fact:trig-orthonormal-basis-cube}	
	$\mathcal{T}$ is an orthonormal basis for $\fnmeas{\intvl^d}$.	
\end{fact} 	

\begin{proof}
	First, we make use of the well-known fact that the constant $1$ function, along with $z \mapsto \sqrt{2}\sin(\pi kz)$ and $z \mapsto \sqrt{2}\cos(\pi kz)$ for all $k \in \Z^+$, collectively form an orthonormal basis for $\fnmeas{\intvl}$.
	(For details, see \citealp{dym1972fourier}.) 
	Thus, the $d$-fold Cartesian product of this collection is an orthonormal basis for $\fnmeas{\intvl^d}$.\footnote{
	This is also an orthonormal basis, so we could similarly represent functions in $\fnmeas{\intvl^d}$ as linear combinations of the elements of this basis and apply the properties of Fact~\ref{fact:parseval-plancherel}.
	However, this representation is unhelpful for our analysis because its elements have large Lipschitz constants and are not ridge functions.}
  Each function in this basis is a product of $d$ functions---one per variable, and each being either a constant, sine, or cosine as above---and can be rewritten as a linear combination of functions from $\mathcal{T}$ using basic product-to-sum trigonometric identities.
	Thus, $\Span{\mathcal{T}} = \fnmeas{\intvl^d}$. 
	
	To complete our proof, it remains to show that all elements of $\mathcal{T}$ are orthogonal and have unit norm. 
	It suffices to show that $\ipmeasl{T_K}{T_{K'}}{\intvl^d} = \indicatorl{K = K'}$ for all $K, K' \in \mathbb{Z}^d$. 
	There are six possible scenarios for this claim depending on which partitioning subsets of $\Z^d$  contain $K$ and $K'$: \emph{(1)} $K, K' \in \cosset$; \emph{(2)} $K, K' \in \sinset$; \emph{(3)} $K = K' = \vec{0}$; \emph{(4)} $K \in \cosset, K' = \vec{0}$ or $K =  \vec{0}, K'\in \cosset$; \emph{(5)} $K \in \sinset, K' = \vec{0}$ or $K =  \vec{0}, K'\in \sinset$; and \emph{(6)} $K \in \sinset, K' \in \cosset$ or $K \in \cosset, K' \in \sinset$. 
	For the sake of simplicity, we only explicitly prove the claim for scenario \emph{(1)}.
	The other cases can be proved with similar trigonometric arguments, all of which involve applying Fact \ref{fact:multi-integral-cosine}.
  For scenario \emph{(1)}, we observe that
	\begin{align*}	
		\ipmeas{T_K}{T_{K'}}{\intvl^d}	
		&= \frac{1}{2^d}\int_{[-1, 1]^d} 2 \cos\paren{\pi \innerprod{K}{x}} \cos\paren{\pi \innerprod{K'}{x}} \dif x \\	
		&= \frac{1}{2^d}\int_{[-1, 1]^d} \bracket{ \cos\paren{\pi \innerprod{K-K'}{x}} -  \cos\paren{\pi \innerprod{K+K'}{x}} }\dif x \\	
		&= \frac{1}{2^d} \bracket{2^d \indicator{K - K' = 0} - 2^d\indicator{K + K' = 0}} \\	
		&= \indicator{K = K'}.
	\end{align*} 		
The last equality holds because if $K + K' = 0$, then either $K$ or $K'$ must belong to $\sinset$ by the definitions of $\sinset$ and $\cosset$.
\end{proof}
We additionally derive the following useful fact about the partial derivatives of elements of the trigonometric basis $\mathcal{T}$.
\begin{fact}[Orthogonality of derivatives of $\mathcal{T}$]\label{fact:trig-pderiv}
	For all $M \in \N^d$ and for all $K, K' \in \Z^d$, 
	\[ \ipmeas{\dderiv{M}{T_K}}{\dderiv{M}{T_{K'}}}{\intvl^d}  = \indicator{K = K'} \pi^{2\abs{M}} K^{2M} .\]	
\end{fact}
\begin{proof}
	The partial derivatives of $T_K$ for every $\mathcal{K} \in \Z^d$ can be exactly characterized by inductively taking derivatives of $\sin$ and $\cos$ functions:
	\begin{equation} \label{eq:deriv-of-trig}\dderiv{M}{T_K}(x) = \begin{cases}
		\pi^{\abs{M}}  T_{K}(x) K^M & \abs{M} \equiv 0 \pmod{4} \\
		\pi^{\abs{M}}  T_{-K}(x) K^M&   \abs{M} \equiv 1 \pmod{4}\  \& \ K \in \sinset\\
		-\pi^{\abs{M}}  T_{-K}(x) K^M &  \abs{M} \equiv 1 \pmod{4}\  \& \ K \in \cosset \cup \setl{\vec{0}} \\
		-\pi^{\abs{M}}  T_{K}(x) K^M &  \abs{M} \equiv 2 \pmod{4} \\
		-\pi^{\abs{M}}  T_{-K}(x) K^M &  \abs{M} \equiv 3 \pmod{4}\  \& \ K\in \sinset\\
		\pi^{\abs{M}}  T_{-K}(x) K^M &  \abs{M} \equiv 3 \pmod{4}\  \& \ K \in \cosset \cup \setl{\vec{0}}.
	\end{cases} 
	\end{equation}
	The conclusion follows by applying the orthonormality of trigonometric basis elements from Fact \ref{fact:trig-orthonormal-basis-cube} to Equation~\eqref{eq:deriv-of-trig}. 
\end{proof}

To prove that a function $f \in \fnmeas{\intvl^d}$ can be represented by a linear combination of sufficiently many random ReLUs, we first show that $f$ can be approximated by a low-degree trigonometric polynomial.
To do so, we upper-bound the higher-order coefficients of the trigonometric expansion of $f$. 
Obtaining these bounds requires taking partial derivatives of $f$ by differentiating term-by-term the trigonometric expansion of $f$. 
However, this is not always possible; for instance, if $f(x) = x_1$, the terms of the trigonometric expansion of $\pderivl{f}{x_1}$ do not correspond to the term-by-term derivatives of the expansion of $f$.\footnote{
Because $\pderivl{f}{x_1} = 1$, its trigonometric expansion $\pderivl{f}{x_1} = \sum_{K \in \Z^d} \beta_K T_K$ will have $\beta_K = \indicatorl{K = \vec{0}}$. Because $f = \sum_{K \in \Z^d} \alpha_K T_K$ will have $\alpha_{K} \neq 0$ for some $K \neq \vec{0}$, $\beta_K \neq 0$ if term-by-term differentiation were possible. Since this contradicts the expansion of $\pderivl{f}{x_1}$, term-by-term differentiation is impossible in this case.
}
We define a notion of \emph{boundary periodicity} that lets us perform term-by-term differentiation:

\begin{definition}[Periodic boundary conditions]
	$f \in \fnmeas{\intvl^d}$ satisfies the \emph{periodic boundary conditions} if for all $i \in [d]$ and for all $x \in \intvl^d$
	\[f(x_1, \dots, x_{i-1}, -1, x_{i+1}, \dots, x_{d}) = f(x_1, \dots, x_{i-1}, 1, x_{i+1}, \dots, x_{d}).
	\]
\end{definition}
Note that all basis elements in $\mathcal{T}$ satisfy the periodic boundary conditions. 
The next lemma gives sufficient conditions for term-by-term differentiation of a function's trigonometric representation.

\begin{lemma}[Term-by-term differentiation of trigonometric basis representations]\label{lemma:trig-expansion-deriv}
  \sloppy
	Consider some $f \in \fnmeas{\intvl^d}$ and $i \in [d]$ such that $f$ satisfies the periodic boundary conditions, $f$ is differentiable with respect to $x_i$, and $\pderivl{f}{x_i} \in  \fnmeas{\intvl^d}$. Then, $f$ and $\pderivl{f}{x_i}$  have trigonometric expansions of the form%
  \[f = \sum_{K \in \Z^d} \alpha_K T_K \qquad \& \qquad  \pderiv{f}{x_i} = \sum_{K \in \Z^d} \beta_K T_K,\]
	where their coefficients $(\alpha_K)_{K \in \Z^d}, (\beta_K)_{K \in \Z^d} $  are related as follows:
	\begin{equation}\label{line:k}
		\beta_K = \begin{cases}
		\pi K_i \alpha_{-K} & K \in \cosset  \\
		- \pi K_i \alpha_{-K} & K \in \sinset \\
		0 & K = \vec{0} .
		\end{cases}
	\end{equation}
	Therefore,
	\[\pderiv{f}{x_i} = \sum_{K \in \Z^d} \alpha_K \pderiv{T_{K}}{x_i}.\]
  \fussy
\end{lemma}
\begin{proof}
Without loss of generality, let $i = 1$.
Because each of $f$ and $\pderivl{f}{x_1}$ is in $\fnmeas{\intvl^d}$, there exist $\alpha$ and $\beta$ by Fact \ref{fact:trig-orthonormal-basis-cube} such that $f$ and $\pderivl{f}{x_1}$ are exactly represented by the expansions given in the lemma statement. 
It remains to show that \eqref{line:k} holds.
We fix any $K \in \cosset$, where $T_K(x) = \sqrt{2} \cos(\pi \innerprod{K}{x})$ and $\pderivl{T_K(x)}{x_1} = -\sqrt{2} \pi K_1 \sin(\pi \innerprod{K}{x})$. 
By Fact \ref{fact:parseval-plancherel}, each coefficient of the representation is an inner-product: $\alpha_K = \ipmeasl{f}{T_K}{\intvl^d}$ and $\beta_K = \ipmeasl{\pderivl{f}{x_1}}{T_K}{\intvl^d}$.
Moreover, $\beta_K$ is related to $\alpha_{-K}$, as shown in the following:
\begin{align*}
	\beta_K 
	&= \ipmeas{\pderiv{f}{x_1}}{T_K}{\intvl^d} 
	=  \frac{\sqrt{2}}{2^d} \int_{\intvl^d} \pderiv{f(x)}{x_1} \cos(\pi \innerprod{K}{x}) \dif x \\
	&= \frac{\sqrt{2}}{2^d} \int_{\intvl^{d-1}} \int_{-1}^{1} \pderiv{f(x)}{x_1} \cos(\pi \innerprod{K}{x}) \dif x_1 \dif x_{-1} \\
	&=\frac{\sqrt{2}}{2^d} \int_{\intvl^{d-1}} \bracket{ f(x) \cos(\pi \innerprod{K}{x}) \bigg\lvert_{-1}^{1} + \int_{-1}^{1} f(x) \pi K_1 \sin(\pi \innerprod{K}{x}) \dif x_1 } \dif x_{-1}    \stepcounter{equation}\tag{\theequation}\label{line:o} \\
	&=\frac{\sqrt{2}}{2^d} \int_{\intvl^{d}} f(x) \pi K_1 \sin(\pi \innerprod{K}{x}) \dif x   \stepcounter{equation}\tag{\theequation}\label{line:p} 
	= \pi K_1 \ipmeas{f}{T_{-K}}{\intvl^d} 
	= \pi K_1 \alpha_{-K}.
\end{align*}
We integrate by parts for Equation \eqref{line:o} and take advantage of the periodic boundary conditions of $f$ and $T_K$ for Equation \eqref{line:p}.
A symmetric argument proves the claim for $K \in \sinset$. 
When $K = \vec{0}$, we repeat the above argument, and the periodic boundary conditions of $f$ imply that $\beta_{\vec{0}} = 0$.
\end{proof}

The subspaces of $\fnmeas{\intvl^d}$ of primary interest in our analysis are spanned by a set of orthonormal functions that are indexed by the integer lattice points contained in given Euclidean balls.
The next fact upper- and lower-bounds the number of such points (and hence the dimension of such a subspace).

\begin{fact}[Restatement of Fact \ref{fact:Qkd-ub}]
	For all $d \in \Z^+$ and $k \geq 1$, 
  \[ Q_{k, d} = \exp\paren{\Theta\paren{\min\paren{d \log\paren{\frac{k^2}{d}+2}, k^2 \log\paren{\frac{d}{k^2}+2}}}} . \]
\end{fact}

\begin{proof}
For the upper bound, we use the fact that $\norm[1]{K} \leq \norm[2]{K}^2$ for all $K \in \Z^d$:
	\begin{align*}
		Q_{k,d}
		&= \abs{\set{K \in \Z^d: \norm[2]{K} \leq k}} 
		\leq \abs{\set{K \in \Z^d: \norm[1]{K} \leq k^2}} \\
		&\leq \abs{\set{K \in \N^{2d}: \norm[1]{K} \leq k^2}} \stepcounter{equation}\tag{\theequation}\label{line:d} \\
		&\leq {\ceil{k^2} + 2d - 1 \choose \ceil{k^2}} . \stepcounter{equation}\tag{\theequation}\label{line:e}
	\end{align*}
	Inequality \eqref{line:d} holds because we replace each integer in $K$ from the previous line with two natural numbers (there would be equality if we forced one of each pair of natural numbers to equal zero). 
	Line \eqref{line:e} follows from a standard stars-and-bars counting argument.
	Note that \[{\ceil{k^2} + 2d - 1 \choose \ceil{k^2}} = {\ceil{k^2} + 2d - 1 \choose 2d - 1}.\] We show two separate upper-bounds on that quantity, which together prove the claim:
	\begin{align*}
	Q_{k, d} &\leq {\ceil{k^2} + 2d -1 \choose 2d -1} \leq \paren{\frac{e\ceil{k^2}}{2d-1} + e}^{2d-1} \leq \exp\paren{\Theta\paren{d \log\paren{\frac{k^2}{d}+2}}}; \\
	Q_{k, d} &\leq {\ceil{k^2} + 2d -1 \choose \ceil{k^2}} \leq \paren{\frac{2ed}{\ceil{k^2}}+e}^{\ceil{k^2}} \leq \exp\paren{\Theta\paren{k^2 \log\paren{\frac{d}{k^2}+2}}}.
	\end{align*}

For the lower bound, we observe that
\[\min\paren{d \log\paren{\frac{k^2}{d}+2}, k^2 \log\paren{\frac{d}{k^2}+2}} = 
\begin{cases}
	d \log\paren{\frac{k^2}{d}+2} & \text{if $ k^2 \geq d$} , \\
	k^2 \log\paren{\frac{d}{k^2}+2} & \text{if $k^2 < d$} .
\end{cases} \]
We will lower-bound $Q_{k,d}$ by the appropriate term in each of the two cases, $k^2 \geq d$ and $k^2 < d$.

For the case $k^2 < d$, we lower-bound $Q_{k,d}$ by a sum of binomial coefficients:
\begin{align*}
  Q_{k,d}
  & = \abs{\set{K \in \Z^d: \sum_{i=1}^d K_i^2 \leq k^2}} \\
  & \geq \abs{\set{K \in \bit^d : \sum_{i=1}^d K_i \leq k^2}} \\
  & = \binom{d}{0} + \binom{d}{1} + \dots + \binom{d}{\floor{k^2}} .
\end{align*}
If $\floorl{k^2} \leq d/2$, then the sum of binomial coefficients is at least the last one, which we bound using
\[
  \binom{d}{\floor{k^2}}
  \geq \exp\paren{\floor{k^2} \ln\frac{d}{\floor{k^2}}}
  \geq \exp\paren{\frac{\floor{k^2}}{2} \ln\paren{\frac{d}{\floor{k^2}} + 2}}
  = \exp\paren{\Theta\paren{k^2\ln\paren{\frac{d}{k^2} + 2}}}
  .
\]
Otherwise, if $d/2 < \floorl{k^2} < d$, the sum of binomial coefficients is at least $2^{\floorl{k^2}}$, and
\[
  2^{\floor{k^2}}
  = \exp\paren{ (\ln 2) \floor{k^2} }
  \geq \exp\paren{ \frac{\ln 2}{\ln 4} \floor{k^2} \ln\paren{\frac{d}{\floor{k^2}}+2} }
  = \exp\paren{\Theta\paren{k^2\ln\paren{\frac{d}{k^2} + 2}}}
  .
\]

When $k^2 \geq d$, we show that $Q_{k,d}$ grows at a rate similar to that of the volume of a $d$-dimensional ball of sufficiently large radius $\Theta(k)$.
To do so, we regard each $K \in \mathcal{K}_{k,d}$ as an element of $\R^d$, and define
\[
  A_{k,d} := \set{x \in \R^d: \ \min_{K \in \mathcal{K}_{k,d}} \norm[\infty]{x - K} \leq \frac{1}{2}} .
\]
This is the Minkowski sum of $\mathcal{K}_{k,d}$ and the $\ell_\infty$ ball of radius $1/2$ in $\R^d$.
Note that $A_{k,d}$ has Lebesgue measure $\vol(A_{k,d}) = \absl{\mathcal{K}_{k,d}} = Q_{k,d}.$
Let $B_2^d(r) := \setl{x \in \R^d: \norm[2]{x} \leq r}$ be the $d$-dimensional Euclidean ball of radius $r$. 
We claim that $B_2^d(k - \sqrt{d}/2) \subset A_{k,d}$, which in turn implies
\[ Q_{k,d} \geq \vol\paren{B_2^d\paren{k - \sqrt{d}/2}} . \]
To see why this claim holds, consider any $x \in B_2^d(k - \sqrt{d}/2)$.
We'll show that $x \in A_{k,d}$.
Indeed, there exists some $y \in \Z^d$ such that $\norm[\infty]{x - y} \leq 1/2$, and hence this $y$ also satisfies $\norm[2]{x - y} \leq \sqrt{d}/2$.
By the triangle inequality,
\begin{align*}
  \norm[2]{y}
  & \leq \norm[2]{x} + \norm[2]{x-y} \\
  & \leq \paren{k - \frac{\sqrt{d}}2} + \frac{\sqrt{d}}2
  = k .
\end{align*}
Thus, $y \in \mathcal{K}_{k,d}$, which implies $x \in A_{k,d}$.

To complete our lower-bound on $Q_{k,d}$, we observe that 
\begin{align*}
	Q_{k,d}
	&\geq \vol\paren{B_{d}\paren{k - \frac{1}{2}\sqrt{d}}}
	\geq \vol\paren{B_{d}\paren{\frac{k}{2}}} \\
	&= \frac{\pi^{d/2} (k/2)^{d}}{\Gamma\paren{\frac{d}{2} + 1}} 
	\geq \paren{\frac{\pi k^2}{2d + 4}}^{d/2}
	\geq \exp\paren{\Theta\paren{d \log\paren{\frac{k^2}{d}+2}}},
\end{align*}
where $\Gamma$ is the gamma function and we have used a standard bound on the volume of the $d$-dimensional Euclidean ball.
\end{proof}

\section{Supporting lemmas for upper-bounds for Lipschitz functions}\label{section:proofs-Lipschitz}

This appendix supports Section \ref{sec:ub-cube-lipschitz}, which presents and proves Theorem \ref{thm:ub-cube-lipschitz}, the main upper-bound on the minimum width RBL network needed to approximate a Lipschitz function. 
It contains the proofs of the key Lemmas \ref{lemma:trig-lipschitz-approx} and \ref{lemma:ub-cube-trig-bounded-degree}, which are given in Appendices \ref{ssec:trig-lipschitz-approx} and \ref{ssec:ub-cube-trig-bounded-degree-app} respectively.

\subsection{Trigonometric polynomial approximation for Lipschitz functions}\label{ssec:trig-lipschitz-approx}

\begin{lemma}[Restatement of Lemma~\ref{lemma:trig-lipschitz-approx}]\label{lemma:trig-lipschitz-approx-restate}
	Fix some $L, \eps > 0$ with $\frac{L}{\eps} \geq 1$ and consider any function $f \in L^2([-1, 1]^d)$ with $\norm[\lip]{f} \leq L$ and $\abs{\EE[\bx]{f(\bx)}} \leq L$. 
	Then, taking $k = {\frac{L}{\eps}}$, there exists a bounded-degree trigonometric polynomial \[P(x) = \sum_{K \in \mathcal{K}_{k,d}} \beta_K T_{K}\paren{\frac{x}{2}} \]
	such that 
	$\norm[\intvl^d]{f - P} \leq \eps.$
	Moreover, $\abs{\beta_K} \leq L$ for all $K$.\end{lemma}

\begin{proofnoqed}[Proof of Lemma~\ref{lemma:trig-lipschitz-approx-restate}]
	To give a low-degree trigonometric polynomial approximation for $f$, we transform $f$ into a function $\tilde{f}$ that satisfies periodic boundary conditions, apply Lemma~\ref{lemma:trig-lipschitz-approx-periodic} to approximate $\tilde{f}$ with trigonometric polynomial $\tilde{P}$, and obtain $P$ from $\tilde{P}$. 	
	Roughly, the argument proceeds as follows:
	\begin{enumerate}
		\item\label{step:1} We define $\bar{f}: \intv^d \to \R$ to be a rescaling and shift of $f$ so that its domain is the cube $\intv^d$. 
		That is, for $x \in \intvl^d$ and $y \in \intv^d$, $\bar{f}(y) = f(2y - \vec{1})$ and $f(x) = \bar{f}\parenl{(x + \vec{1})/2}$. %
		Then it holds that $\norml[\lip]{\bar{f}} \leq 2L$ and $\absl{\EEl[\by \sim\intv^d]{\bar{f}(\by)}} = \absl{\EEl[\bx\sim \intvl^d]{{f}(\bx)}} \leq L$. 
		\item\label{step:2} We define $\tilde{f}: \intvl^d \to \R$ by reflecting $\bar{f}$ across orthants as follows: $\tilde{f}(x) = \bar{f}(\signl{x} \odot x)$, where $\sign{x} := (\signl{x_1}, \dots, \signl{x_d})$ and $\odot$ represents element-wise multiplication. 
		The function $\tilde{f}$ is $2L$-Lipschitz, satisfies the periodic boundary conditions, and has
    \[ \abs{\EE[\bx\sim \intvl^d]{\tilde{f}(\bx)}} = \abs{\EE[\by\sim \intv^d]{\bar{f}(\by)}} \leq L . \]
		\item\label{step:3} We find a low-degree trigonometric polynomial $\tilde{P}$ that $\eps$-approximates $\tilde{f}$ over $\intvl^d$.
		\item\label{step:4} Such a $\tilde{P}$ must $\eps$-approximate $\tilde{f}$ in at least one of the $2^d$ unit cubes contained in the orthants of $\intvl^d$. Therefore, there exists some sign vector $\nu \in \flip^d$ such that $\bar{f}(y)$ is approximated by $\tilde{P}(\nu \odot y)$ on $\intv^d$.
		\item\label{step:5} By shifting and rescaling $\tilde{P}(\nu \odot y)$, we obtain a trigonometric polynomial $P$ that $\eps$-approximates $f$ on $\intvl^d$ as desired.
	\end{enumerate}

    \begin{figure}[h!]
    \centering
    \includegraphics[width=0.8\textwidth]{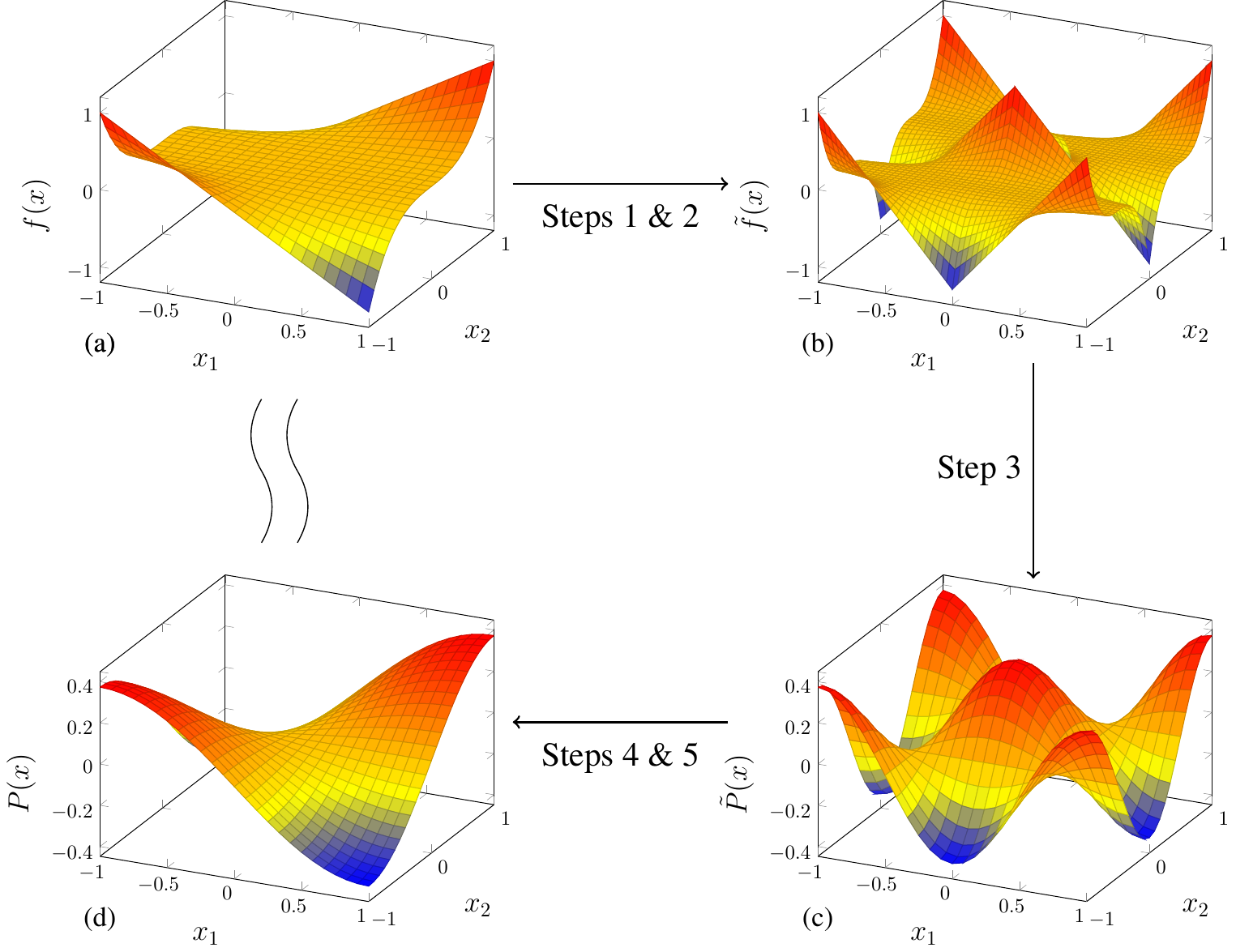}
    	\ignore[true]{
		\begin{tikzpicture}[scale=0.9]  		
		\begin{scope}[scale=0.7]
 			\begin{axis}[domain=-1:1,y domain=-1:1, xlabel={$x_1$}, ylabel={$x_2$},zlabel={$f(x)$}]				
    				\addplot3[surf] {x * y^3};
  			\end{axis}
  		\end{scope}
  		\begin{scope}[xshift=8cm,scale=0.7]
 			\begin{axis}[domain=-1:1,y domain=-1:1, xlabel={$x_1$}, ylabel={$x_2$},zlabel={$\tilde{f}(x)$}]	
    				\addplot3[surf, domain=0:1, y domain=0:1] {(2*x - 1) * (2 * y - 1)^3};
				\addplot3[surf, domain=-1:0, y domain=0:1] {(-2*x - 1) * (2 * y - 1)^3};
				\addplot3[surf, domain=0:1, y domain=-1:0] {(2*x - 1) * (-2 * y - 1)^3};
				\addplot3[surf, domain=-1:0, y domain=-1:0] {(-2*x - 1) * (-2 * y - 1)^3};
  			\end{axis}
  		\end{scope}
  		\begin{scope}[xshift=8cm,yshift=-6cm,scale=0.7]
 			\begin{axis}[domain=-1:1,y domain=-1:1, xlabel={$x_1$}, ylabel={$x_2$},zlabel={$\tilde{P}(x)$}]	
    				\addplot3[surf] {0.132 *sqrt(2) * cos(pi * (deg(x)+deg(y)))+0.132 *sqrt(2) * cos(pi * (deg(x)-deg(y)))};
  			\end{axis}
  		\end{scope}
  		\begin{scope}[xshift=0cm,yshift=-6cm,scale=0.7]
 			\begin{axis}[domain=-1:1,y domain=-1:1, xlabel={$x_1$}, ylabel={$x_2$},zlabel={$P(x)$}]	
    				\addplot3[surf] {0.132 *sqrt(2) * cos(pi/2 * (deg(x+1)+deg(y+1)))+0.132 *sqrt(2) * cos(pi/2 * (deg(x)-deg(y)))};
  			\end{axis}
  		\end{scope}
  		\draw[thick,->] (5,2) to (7,2);
  		\draw (6,1.5) node {\footnotesize Steps 1 \& 2};
  		\draw (0,-0.25) node {(a)};
  		\draw (8,-0.25) node {(b)};
  		\draw (8,-6.25) node {(c)};
  		\draw (0,-6.25) node {(d)};
  		\draw[thick,->] (10,0) to (10,-2);
 		\draw (9.25,-1) node {\footnotesize Step 3};
  		\draw[thick,<-] (5,-4) to (7,-4);
		\draw (6,-4.5) node {\footnotesize Steps 4 \& 5};
  		\def\ytranspose{-0.2};
  		\draw (2.75,-0.4+\ytranspose) to[bend right] (2.75,-1+\ytranspose);
  		\draw (2.75,-1+\ytranspose) to[bend left]  (2.75,-1.6+\ytranspose);
  		\draw (2.25,-0.4+\ytranspose) to[bend right] (2.25,-1+\ytranspose);
  		\draw (2.25,-1+\ytranspose) to[bend left]  (2.25,-1.6+\ytranspose);
		\end{tikzpicture}
		}
    \caption{A depiction of the function transformations used to give an approximation of $f$ in Lemma~\ref{lemma:trig-lipschitz-approx}. The original function $f$ is in (a), which is scaled and reflected to yield a function $\tilde{f}$ with periodic boundary conditions in (b), which is given a trigonometric polynomial approximation $\tilde{P}$ in (c), which is in turn scaled and shifted to obtain $P$ approximating the original $f$ in (d).}
\end{figure}
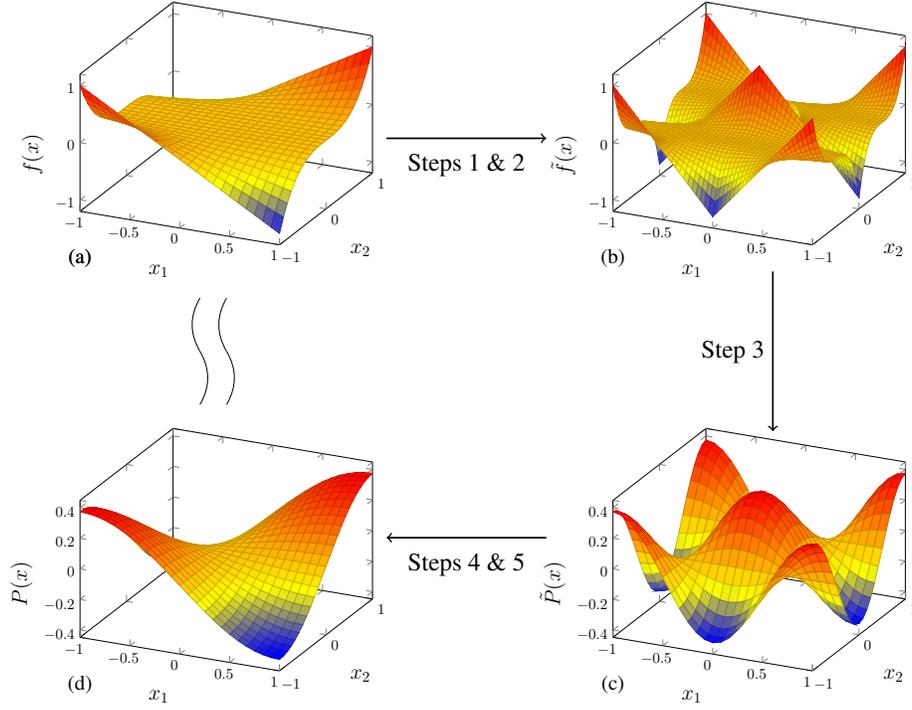

	Steps (\ref{step:1}) and (\ref{step:2}) are immediate. 
	
	Step (\ref{step:3}) follows from Lemma~\ref{lemma:trig-lipschitz-approx-periodic}.
	Because $\tilde{f}$ is $2L$-Lipschitz, $\tilde{f}$ satisfies the periodic boundary conditions, $\absl{\EEl[\bx\sim \intvl^d]{\tilde{f}(\bx)}} \leq L$, and $\fracl{2L}{\eps}\geq 2$, Lemma~\ref{lemma:trig-lipschitz-approx-periodic} guarantees the existence of some trigonometric polynomial 
	\[\tilde{P}(x) = \sum_{K \in \mathcal{K}_{k,d}} \tilde{\beta}_K T_K(x)\]
	such that $\norml[\intvl^d]{\tilde{f} - \tilde{P}} \leq \eps$ and $\absl{\tilde{\beta}_K} \leq L$ for all $K$.

	For step (\ref{step:4}), if $\tilde{P}$ is an $\eps$-approximator for $\tilde{f}$ over $\fnmeas{\intvl^d}$, then there must exist a unit cube in some orthant corresponding to some $\nu \in \flip^d$ where $\tilde{P}$ also $\eps$-approximates $\tilde{f}$. That is,
	\[\EE[\by \sim \intv^d]{\paren{\tilde{P}(\nu \odot \by) - \bar{f}( \by)}^2} \leq \eps^2.\]
	
	For step (\ref{step:5}), by translating the distribution from $\intvl^d$ to $\intv^d$ and taking $P(x) := \tilde{P}\parenl{ \nu \odot (x + \vec{1})/2} $, we obtain
	\begin{align*}
		\EE[\bx \sim \intvl^d]{\paren{P(\bx) - f(\bx)}^2} = \EE[\by \sim \intv^d]{\paren{\tilde{P}(\nu \odot \by) - \bar{f}( \by)}^2} 		
	\end{align*} 
	
	It remains to show that we can represent $P$ as a proper trigonometric polynomial with halved frequencies and bounded coefficients. 
	We do so by examining each term of the expansion of $\tilde{P}$. Fix any $K \in \Z^d$ with $\norm[2]{K} \leq k$ and $K \in \sinset$. 
	Then, $T_K(y) = \sqrt{2} \sin(\pi \innerprodl{K}{y})$. 
	Consider the term corresponding to $K$ of $P(x)$ represented as an expansion of $\tilde{P}$, $\tilde{\beta}_K T_K\parenl{  \nu \odot (x + \vec{1})/2}$. By rearranging its inner product and applying sum-of-angles trigonometric identities, we obtain the following identity:
	\begin{align*}
		T_K\paren{ \half \nu \odot (x + \vec{1})} 
		&= \sqrt{2}  \sin\paren{\frac{\pi}{2} \innerprodl{\nu \odot K}{x} + \frac{\pi}{2} \innerprodl{\nu \odot K}{\vec{1}}} \\
		&= \begin{cases}
			\sqrt{2}  \sin\paren{\frac{\pi}{2} \innerprod{\nu \odot K}{x} }&\innerprodl{\nu \odot K}{\vec{1}} \equiv 0 \pmod 4 \\
			\sqrt{2}  \cos\paren{\frac{\pi}{2} \innerprod{\nu \odot K}{x} }&\innerprodl{\nu \odot K}{\vec{1}} \equiv 1 \pmod 4 \\
			-\sqrt{2}  \sin\paren{\frac{\pi}{2} \innerprod{\nu \odot K}{x} }&\innerprodl{\nu \odot K}{\vec{1}} \equiv 2 \pmod 4 \\
			-\sqrt{2}  \cos\paren{\frac{\pi}{2} \innerprod{\nu \odot K}{x} } & \innerprodl{\nu \odot K}{\vec{1}} \equiv 3 \pmod 4 .
		\end{cases}
			\end{align*}
      This yields
  the final representation for $T_K$ functions:
	\begin{align*}
			T_K\paren{ \half \nu \odot (x + \vec{1})} 
	&= \begin{cases}
			T_{\nu \odot K}(\frac{x}{2})&\innerprodl{\nu \odot K}{\vec{1}} \equiv 0 \pmod 4 \\
			T_{-\nu \odot K}(\frac{x}{2})&\innerprodl{\nu \odot K}{\vec{1}} \equiv 1 \pmod 4 \\
			-T_{\nu \odot K}(\frac{x}{2})&\innerprodl{\nu \odot K}{\vec{1}} \equiv 2 \pmod 4 \\
			-T_{-\nu \odot K}(\frac{x}{2})&\innerprodl{\nu \odot K}{\vec{1}} \equiv 3 \pmod 4 .
		\end{cases}
	\end{align*}
	Similarly,
	\[T_{-K}\paren{ \half \nu \odot (x + \vec{1})} = 
	\begin{cases}
		T_{-\nu \odot K}(\frac{x}{2})&\innerprodl{\nu \odot K}{\vec{1}} \equiv 0 \pmod 4 \\
		-T_{\nu \odot K}(\frac{x}{2})&\innerprodl{\nu \odot K}{\vec{1}} \equiv 1 \pmod 4 \\
		-T_{-\nu \odot K}(\frac{x}{2})&\innerprodl{\nu \odot K}{\vec{1}} \equiv 2 \pmod 4 \\
		T_{\nu \odot K}(\frac{x}{2})&\innerprodl{\nu \odot K}{\vec{1}} \equiv 3 \pmod 4  .
	\end{cases}\]
	
	Using these identities, we can rewrite $P$ as its own trigonometric polynomial with coefficients $\beta_K$ for all $K \in \Z^d$ such that $\beta_K \in \setl{\tilde{\beta}_{\nu \odot K}, - \tilde{\beta}_{\nu \odot K}}$ if $\innerprodl{\nu \odot K}{\vec{1}} \equiv 0 \pmod 2$, and $\beta_K \in \setl{\tilde{\beta}_{-\nu \odot K}, - \tilde{\beta}_{-\nu \odot K}}$ otherwise. 
	Due to the existence of such $\beta_K$ coefficients, the following trigonometric polynomial approximates $f$ over $[-1, 1]^d$:
	\[P(x) = \sum_{K \in \mathcal{K}_{k,d}} \tilde{\beta}_K T_K\paren{\frac{1}{2} \nu \odot (x + \vec{1})} = \sum_{K \in \mathcal{K}_{k,d}} \beta_K T_{K}\paren{\frac{x}{2}}. \qedhere \]
\end{proofnoqed}

\subsection{RBL ReLU network approximation for trigonometric polynomials}\label{ssec:ub-cube-trig-bounded-degree-app}

In this section, we give a general purpose lemma that bounds the width needed to approximate trigonometric polynomials of bounded degree. 

\begin{lemma}[Restatement of Lemma~\ref{lemma:ub-cube-trig-bounded-degree}]\label{lemma:ub-cube-trig-bounded-degree-restate}
	Fix some $\delta \in (0, \halfl]$, $\eps > 0$, $\rho \in (0, 1]$, $k \geq 1$, and $d \in \mathbb{Z}^+$. 
	Then, there exists some symmetric ReLU parameter distribution $\mathcal{D}_k$ such that 
	for any trigonometric polynomial \[P(x) = \sum_{K \in \mathcal{K}_{k,d}}\beta_K T_K(\rho x)\] with $\abs{\beta_K} \leq \betam$ for all $K \in \mathcal{K}_{k,d}$, 
	\[\minwidth{P}{\eps}{\delta}{\intvl^d}{\mathcal{D}_k} \leq O\paren{\frac{ \betam^2 d^2 k^4 }{\eps^2} Q_{k,d}^2  \ln \paren{\frac{1}{\delta}}}.\]

\end{lemma}

We first define the specific symmetric ReLU parameter distribution $\mathcal{D}_k$ used in the proof, which can be shown to meet the symmetry criteria spelled out in Definition \ref{def:sym-relu-param-dist}. (As a result, the lower-bounds on the minimum width in Theorems \ref{thm:lb-cube-nonexplicit} and \ref{thm:lb-cube-sine} hold for $\dprodk$.)

\begin{definition}[Symmetric ReLU parameter distribution $\mathcal{D}_k$ for $\intvl^d$ upper-bounds]\label{def:Dk}
	Define $\dprod_k := \dbias \times \dweightk$ as a product distribution with the following components:
	\begin{itemize}
		\item $\dbias$ is the uniform distribution over $[-2\sqrt{d}, 2\sqrt{d}]$; and
		\item $\dweightk$ is a distribution over weights $\bw$ taking value in $\sph$. To draw $\bw$ from $\dweightk$, draw $\bK$ uniformly at random from $\mathcal{K}_{k,d}$ and let $\bw := \fracl{\bK}{\norm[2]{K}}$. (If $\bK = \vec{0}$, let $\bw := \fracl{\vec{1}}{\sqrt{d}}$.)
	\end{itemize}
\end{definition}
We also introduce notation to represent the set of vectors contained in $\mathcal{K}_{k,d}$ that generate each $w \in \supp(\dweightk) \subset \sph$:
\[\mathcal{K}_{k,d,w} := \begin{cases}
	\set{K \in \mathcal{K}_{k,d}: K = \eta w, \eta \geq 0} & w = \frac{1}{\sqrt{d}} \vec{1} \\
	\set{K \in \mathcal{K}_{k,d}: K = \eta w, \eta > 0} & \text{otherwise} .
	\end{cases} \]
Note that every $w \in \supp(\dweightk)$ is drawn with probability $\fracl{\absl{\mathcal{K}_{k,d,w}}}{Q_{k,d} }$, which is at least $\fracl{1}{Q_{k, d} }$ and at most $\fracl{(k+1)}{Q_{k,d} }$.

To prove Lemma \ref{lemma:ub-cube-trig-bounded-degree}, we represent $P$ as an expectation over random ReLU features with parameters drawn from $\mathcal{D}_k$.
We first express each trigonometric basis element $T_K$ as an expectation over random ReLUs.
We leverage the fact that each individual $T_K$ is a ridge function (that is, $T_K(x) = \phi(\innerprodl{K}{x})$ for some $\phi$).
In the following lemma, we show that every ridge function on $\intvl^d$ can be represented as a mixture of ReLUs with random bias terms $\bb$ drawn from $\dbias$.

\begin{lemma}[Representing ridge functions as a mixture of ReLUs]\label{lemma:relu-univariate-representation}
	Let $\phi: [-\sqrt{d}, \sqrt{d}] \to \R$ be twice differentiable and let $f: \intvl^d \to \R$ be $f(x) = \phi(\innerprod{v}{x})$ for some $v \in \sph$. 
	Then, for all $x \in [-1, 1]^d$,
	\[f(x) = \EE[\bb \sim \dbias]{\psi(\bb) \relu\paren{\innerprod{v}{x} - \bb}},\]
	where
	\[\psi(b) := 
	\begin{cases}
		4 \sqrt{d} a_0 := \frac{16}{\sqrt{d}} \phi(-\sqrt{d}) - 4 \phi'(-\sqrt{d}) & b \in [-2\sqrt{d}, -\frac{3}{2} \sqrt{d}) \\
		4 \sqrt{d} a_1 := -\frac{16}{\sqrt{d}} \phi(-\sqrt{d}) + 12 \phi'(-\sqrt{d}) & b \in [-\frac{3}{2} \sqrt{d}, -\sqrt{d}) \\
		4 \sqrt{d} \phi''(b) & b \in [-\sqrt{d}, \sqrt{d}] \\
		0 & b \in (\sqrt{d}, 2\sqrt{d}] .
	\end{cases}\]
\end{lemma}
\begin{proofnoqed}
	We expand the expectation over $\bb$. For $x \in \intvl^d$, let $z := \innerprod{v}{x} \in [-\sqrt{d}, \sqrt{d}]$. We have the following:
	\begin{align*}
		 &\EE[\bb \sim \dbias]{\psi(\bb) \relu\paren{\innerprod{v}{x} - \bb}}\\
		 &~~~~= a_0 \int_{-2\sqrt{d}}^{-\frac{3}{2} \sqrt{d}} \relu(z - b)\dif b  + a_1 \int_{-\frac{3}{2}\sqrt{d}}^{-\sqrt{d}} \relu(z - b)\dif b + \int_{-\sqrt{d}}^{\sqrt{d}} \phi''(b) \relu(z - b) \dif b \\
		 &~~~~= a_0 \paren{zb - \frac{1}{2} b^2} \bigg\rvert_{-2\sqrt{d}}^{-\frac{3}{2} \sqrt{d}} + a_1 \paren{zb - \frac{1}{2} b^2} \bigg\rvert_{-\frac{3}{2}\sqrt{d}}^{-\sqrt{d}} +  \int_{-\sqrt{d}}^{z} \phi''(b) (z - b) \dif b \\
		 &~~~~= \frac{\sqrt{d}}{2}z \paren{a_0  + a_1} + \frac{d}{8} \paren{7 a_0 + 5 a_1} + \paren{\phi'(b) (z - b)}\bigg\rvert_{-\sqrt{d}}^z - \int_{-\sqrt{d}}^z  \phi'(b) \cdot (-1) \dif b \\
		 &~~~~=  z\phi'(-\sqrt{d}) + \phi(-\sqrt{d}) + \sqrt{d} \phi'(-\sqrt{d})  - \phi'(-\sqrt{d})(z + \sqrt{d}) + \phi(z) - \phi(-\sqrt{d}) \\
		 &~~~~= \phi(z) = f(x).
     \qedhere
	\end{align*}
\end{proofnoqed} 

	Once $P$ is represented as an expectation over random ReLUs with parameters drawn from $\mathcal{D}_k$, we conclude the proof by arguing that this expectation can be closely approximated with high probability by a linear combination of sufficiently many randomly sampled ReLUs.
  We do so by applying a concentration bound due to \cite{yurinskiui1976exponential} for sums of independent random variables taking values in a Hilbert space.
  We use a convenient version of the bound from \citet[Lemma 4]{rr08}:
	\begin{lemma}[Concentration inequality for Hilbert spaces]\label{lemma:rr08}
		Let $\bhi{1}, \dots, \bhi{r}$ be independent random variables that take values in a Hilbert space with norm $\norm{\cdot}$ such that $\norml{\bhi{i}} \leq m$ for all $i$. Then, for any $\delta \in (0, 1)$, with probability at least $1 - \delta$,
		\[\norm{\frac{1}{r} \sum_{i=1}^r \bhi{i} - \EE{\frac{1}{r} \sum_{i=1}^r \bhi{i}}} \leq \frac{m}{\sqrt{r}}\paren{1 + \sqrt{2 \log\paren{\frac{1}{\delta}}}}.\]
	\end{lemma}

We are now prepared to formally prove Lemma \ref{lemma:ub-cube-trig-bounded-degree}.

\begin{proof}[Proof of Lemma \ref{lemma:ub-cube-trig-bounded-degree}]
	We first represent any trigonometric monomial $T_K$ as an expected value over weighted ReLUs of the form $\relu\parenl{\innerprod{K / \norm[2]{K}}{ x} + \bb}$ for $\bb \sim \dbias$. 	
	For each $K$, we have $T_K(\rho x) = \phi_K\parenl{\innerprod{K/\norm[2]{K}}{ x}}$, where 
	\[\phi_K(z) = \begin{cases}
		\sqrt{2} \cos(\pi \rho \norm[2]{K} z) & K \in \cosset \\
		\sqrt{2} \sin(\pi \rho \norm[2]{K} z) & K \in \sinset \\
		1 & K = \vec{0} .
	\end{cases} \]
	By Lemma \ref{lemma:relu-univariate-representation},
\begin{align*}
		T_K(\rho x) &= \EE[\bb \sim \dbias]{\psi_K(b) \relu\paren{\frac{1}{\norm[2]{K}}\innerprod{K}{x} - \bb}},
	\end{align*}	
	where $\psi_K$ is the function defined in  Lemma \ref{lemma:relu-univariate-representation} for $\phi_K$.
	Because $\abs{\phi_K(z)} \leq \sqrt{2}$, $\abs{\phi_K'(z)} \leq \sqrt{2} \pi \rho \norm[2]{K}$, and $\abs{\phi_K''(z)} \leq \sqrt{2} \pi^2 \rho^2 \norm[2]{K}^2$ for all $z$, we can bound $\psi_K$: 
	\begin{align*}
		\abs{\psi_K(z)}
		\leq \max\set{\frac{16}{\sqrt{d}} \cdot \sqrt{2}  + 12 \cdot \sqrt{2} \pi \rho \norm[2]{K}, 4 \sqrt{d} \sqrt{2} \pi^2 \rho^2 \norm[2]{K}^2}
		\leq 60\sqrt{d} \paren{\norm[2]{K}^2 + 1}.
	\end{align*}

	Because any sinusoidal basis element $T_K$ can be expressed as an expectation of random ReLUs and because $P$ is a linear combination of a finite number of those basis elements, we can also represent $P$ as an expectation over ReLUs.
	We define $h: \R \times \sph  \to \R$ as
	\[h(b,w) = \frac{Q_{k,d} }{\abs{\mathcal{K}_{k,d,w}}} \sum_{K \in \mathcal{K}_{k,d,w}} \beta_K  \psi_K(b) =  \frac{1 }{\pr[\bw \sim \dweightk]{\bw = w}} \sum_{K \in \mathcal{K}_{k,d,w}} \beta_K  \psi_K(b),\]
	and represent $P(x)$ as an infinite mixture of ReLU functions weighted by $h$ over all $x \in \intvl^d$. 
	\begin{align*}
    \lefteqn{
      \EE[\bb,\bw]{h(\bb, \bw) \relu\paren{\innerprod{\bw}{x} - \bb} }
    } \\
		&=  \sum_{w \in \supp(\dweightk)} \EE[\bb \sim \dbias]{ \sum_{K \in \mathcal{K}_{k,d,w}} \beta_K \psi_K(\bb) \relu\paren{\innerprod{w}{x} - \bb} } \\
		&=  \sum_{w \in \supp(\dweightk)} \sum_{K \in \mathcal{K}_{k,d,w}} \beta_K \EE[\bb \sim \dbias]{\psi_K(\bb) \relu\paren{\frac{1}{\norm[2]{K}}\innerprod{K}{x} - \bb}} \\
		&= \sum_{K \in \mathcal{K}_{k,d}} \beta_K \EE[\bb \sim \dbias]{\psi_K(\bb) \relu\paren{\frac{1}{\norm[2]{K}}\innerprod{K}{x} - \bb}} \\
		&= \sum_{K \in \mathcal{K}_{k,d}} \beta_K T_K(\rho x) \\
    & = P(x) .
	\end{align*}

  To conclude the proof,
  let $(\bwi{1},\bbi{1}),\dots,(\bbi{r},\bwi{r})$ be independent copies of $(\bw,\bb)$, and define $\bhi{i} \in \fnmeas{\intvl^d}$ for $i=1,\dotsc,r$ by
  \[\bhi{i}(x) := h(\bwi{i}, \bbi{i}) \relu\parenl{\innerprodl{\bwi{i}}{x} - \bbi{i}} .
  \]
  Now we apply Lemma~\ref{lemma:rr08} to the random variables $\bhi{1},\dots,\bhi{r}$.
  Note that $\EEl[\bbi{i}, \bwi{i}]{\bhi{i}(x)} = P(x)$.
  To apply the lemma, we first bound $\norml[\intvl^d]{\bhi{i}}$:
  
	\begin{align*}
		\norm[\intvl^d]{\bhi{i}}
		&\leq \max_{ b \in [-2\sqrt{d}, 2\sqrt{d}],w \in \sph, x \in \intvl^d} \abs{h(b, w)\relu\paren{\innerprod{w}{x} - b}}  \\
		&\leq \paren{ \max_{ b,w, x} \abs{\relu\paren{\innerprod{w}{x} - b}}} \paren{\max_{b,w} \abs{h(b, w)} }  \\
		&=\paren{\max_{w, x} \norm[2]{w} \norm[2]{x} + \max_b \abs{b}} \paren{\max_{b,w}  \frac{Q_{k,d} }{\abs{\mathcal{K}_{k,d,w}}} \abs{\sum_{K \in \mathcal{K}_{k,d,w}} \beta_K \psi_K(b)}}  \\
		&\leq 3 \sqrt{d}Q_{k,d} \max_{w} \frac{1}{\abs{\mathcal{K}_{k,d,w}}} \sum_{K \in  \mathcal{K}_{k,d,w}}\abs{ \beta_K}  \cdot 60\sqrt{d} \paren{\norm[2]{K}^2 +1} \\
		&\leq 360 d Q_{k,d} \betam k^2  .
	\end{align*}
	Therefore, with probability $1 - \delta$,
	\begin{align*}
		\inf_{g \in \Span{\bgi{1}, \dots, \bgi{r}}}\norm[\intvl^d]{P - g}
		&\leq \norm[\intvl^d]{\frac{1}{n} \sum_{i=1}^r \bhi{i} - \EE{\frac{1}{r} \sum_{i=1}^n \bhi{i}}} \\
		&\leq \frac{360 d  \betam k^2 Q_{k,d}}{\sqrt{r}} \paren{1 + \sqrt{2 \ln \frac{1}{\delta}}} \leq \eps,
	\end{align*}
	which holds as long as we choose $r$ with
	\[r \geq \frac{360^2 d^2 \betam^2 k^4 Q_{k,d}^2}{\eps^2} \paren{1 + \sqrt{2 \ln \frac{1}{\delta}}}^2.  \]
	Based on Definiton~\ref{def:minwidth}, this gives the desired upper-bound on MinWidth.
	\end{proof}

\section{Supporting lemmas for lower-bounds for Lipschitz functions}\label{sec:lb-app}

This appendix supports Section \ref{sec:lb} by proving Theorems~\ref{thm:lb-cube-nonexplicit} and~\ref{thm:lb-cube-sine}. 

\subsection{General lower-bounds for random features} \label{ssec:general-lb-app}

In Theorem \ref{thm:randict}, we give the most general form of our lower-bound. 
In this setting, we consider linear combinations of features drawn independently from some distribution over functions (which are not required to be ReLUs or even ridge functions). 
We argue that the span of any $r$ such random functions in $\fnmeas{\mu}$ cannot cover more than $r$ dimensions of that function space and that we therefore cannot approximate most of the members of a family of $N$ orthonormal functions if $N \gg r$. 

If the family of $N$ functions satisfies a suitable notion of symmetry with respect to the random features, then we can additionally argue that each function in that family is equally likely to be inapproximable. 
This makes it possible to construct a single explicit function that cannot be approximated with high probability by linear combinations of random features.
We give the relevant notion of symmetry below:

\begin{definition}[Symmetry of random functions]\label{def:sym-random-fn}
Let $\bg$ be an $\fnmeas{\mu}$-valued random variable for some measure $\mu$.
We say $\bg$ is \emph{symmetric} with respect to the set of functions $\Phi = \setl{\varphi_1,\dotsc,\varphi_N} \subset \fnmeas{\mu}$ if the distribution of $\ipmeas{\bg}{\varphi_i}{\mu}$ is the same for all $i=1,\dotsc,N$.
\end{definition}

In fact, strict orthonormality of the hard functions is not needed for our approach; we introduce a notion of ``average coherence,''

which allows us to quantify how far the family is from being orthogonal and prove lower-bounds that depend on this quantity.  
\begin{definition}[Average coherence]
For any set of functions $\Phi = \setl{\varphi_1,\dotsc,\varphi_N} \subset \fnmeas{\mu}$ with $\norml[\mu]{\varphi_i} = 1$ for all $i=1,\dotsc,N$, its \emph{(average) coherence} is
$\kappa(\Phi) := \sqrt{\sum_{i \neq j} \ipmeasl{\varphi_i}{\varphi_j}{\mu}^2}$.
\end{definition}
We are particularly interested in large collections of functions with low coherence. 
Note that a collection of orthogonal functions has zero coherence.
Our main approximation lower bounds in Theorems~\ref{thm:lb-cube-nonexplicit} and~\ref{thm:lb-cube-sine} are achieved using an orthogonal collection.  
However, our general lower bound (Theorem~\ref{thm:randict}) extends to the case where the family of functions has small (but nonzero) coherence, and indeed this version for families with small coherence is useful in extending our general approach to functions over Gaussian space, as we sketch in Appendix~\ref{asec-gaussian}.

The following general lower bound works for any distribution over random features that meets the above symmetry condition and for any set of ``nearly-orthonormal'' functions that have a bounded average coherence $\kappa$.
It is akin to Theorem~19 of \cite{kms20} although that result does not involve a symmetry notion (and hence does not yield an explicit hard function).

\begin{theorem}[Lower-bound for linear combinations of random features]\label{thm:randict}
  Fix a family of functions $\Phi = \setl{\varphi_1,\dotsc,\varphi_N} \subset \fnmeas{\mu}$ with $\norml[\mu]{\varphi_i}^2 = 1$ for all $i=1,\dotsc,N$.
  Let $\bgi{1},\dotsc,\bgi{r}$ be i.i.d.\ copies of an $\fnmeas{\mu}$-valued random variable.
  Then, there exists some $\varphi_i \in \Phi$ such that
  \begin{equation}\label{eq:randict1}
    \EE[\bgi{1},\dotsc,\bgi{r}]{ \inf_{g \in \spangl} \norm[\mu]{g - \varphi_i}^2 }
    \geq 1 - \frac{r \paren{ 1{+}\kappa(\Phi) }}{N} .
 \end{equation}
  In particular, for any $\alpha \in [0,1]$,
  \begin{equation} \label{eq:randict2}
    \kern-5pt %
    \pr[\bgi{1},\dotsc,\bgi{r}]{ \inf_{g \in \spangl}
    \kern-10pt %
  \norm[\mu]{g - \varphi_i}^2 \geq \alpha \paren{ 1 - \frac{r \paren{ 1{+}\kappa(\Phi) }}{N}} }
    \geq (1{-}\alpha) \paren{ 1 - \frac{r \paren{ 1{+}\kappa(\Phi) }}{N} } .
  \end{equation}
  Moreover, if $\bgi{1},\dotsc,\bgi{r}$ are  symmetric with respect to $\Phi$, then \eqref{eq:randict1} and \eqref{eq:randict2} hold for $i = 1$.
\end{theorem}

We recall two tools that will be used in the proof of Theorem \ref{thm:randict}, namely the Hilbert projection theorem and the Boas-Bellman inequality.

\begin{fact}[Hilbert projection theorem \citep{rudin87}]
\label{fact:hilbert-proj}
For some measure $\mu$ and $\gi{1},\dots,\gi{r} \in \fnmeas{\mu}$, consider the subspace $W = \Spanl{\gi{1},\dots,\gi{r}}$ of $\fnmeas{\mu}$. For any $f \in \fnmeas{\mu}$, it holds that
\begin{equation}
  \inf_{g \in W} \norm[\mu]{g - f}^2
  = \norm[\mu]{\Pi_Wf - f}^2
  = \norm[\mu]{f}^2 - \norm[\mu]{\Pi_Wf}^2 ,
\end{equation}
where $\Pi_W \colon \fnmeas{\mu} \to W$ is the orthogonal projection operator for $W$.
Moreover, the orthogonal projection $\Pi_Wf$ depends on $f$ only through $(\ipmeasl{\gi{1}}{f}{\mu},\dotsc,\ipmeasl{\gi{r}}{f}{\mu})$.
\end{fact}

The following is a generalization of Bessel's inequality due to \cite{boas41} and \cite{bellman44}, specialized to our present context.
\begin{fact}[Boas-Bellman inequality]\label{fact:boas-bellman}
  For any $g, \varphi_1,\dotsc,\varphi_N \in \fnmeas{\mu}$,
  \begin{equation}
    \sum_{i=1}^N \ipmeas{g}{\varphi_i}{\mu}^2
    \leq \norm[\mu]{g}^2 \paren{ \max_{1 \leq i \leq N} \norm[\mu]{\varphi_i}^2 +  \kappa(\set{\varphi_1,\dotsc,\varphi_N})}.
  \end{equation}
\end{fact}

\begin{proof}[Proof of Theorem \ref{thm:randict}]
  By the Hilbert projection theorem (Fact \ref{fact:hilbert-proj}), for all $i \in [N]$ we have that
 \[
    \EE[\bgi{1},\dotsc,\bgi{r}]{ \inf_{g \in \spangl} \norm[\mu]{g - \varphi_i}^2}
    = 1 - \EE[\bgi{1},\dotsc,\bgi{r}]{ \norm[\mu]{\Pi_{\spangl} \varphi_i}^2 } .
  \]
  We now upper-bound the sum of the expected norms of the projections of each function in $\Phi$ onto $\spangl$.
  Let $\bu_1,\dotsc,\bu_{\bd}$ be an orthonormal basis for $\spangl$, where $\bd := \dim \spangl$.
  Then
  \begin{align*}
    \sum_{i=1}^N \norm[\mu]{\Pi_{\spangl} \varphi_i}^2
    & = \sum_{i=1}^N \sum_{k=1}^{\bd} \ipmeas{\bu_k}{\varphi_i}{\mu}^2
    = \sum_{k=1}^{\bd} \sum_{i=1}^N \ipmeas{\bu_k}{\varphi_i}{\mu}^2 
    && \text{(Plancherel's identity, Fact~\ref{fact:parseval-plancherel})} \\
    &  \leq \sum_{k=1}^{\bd} \paren{ 1 + \kappa(\Phi) } = \bd \cdot \paren{ 1 + \kappa(\Phi) } 
    && \text{(Fact \ref{fact:boas-bellman})} \\
    & \leq r \cdot \paren{ 1 + \kappa(\Phi) }
    && \text{($\dim \spangl \leq r$)} .
  \end{align*}
  Hence, we conclude by linearity of expectation that 
   \begin{equation}\label{eq:randict3}
   \frac{1}{N} \sum_{i=1}^N \EE[\bgi{1},\dotsc,\bgi{r}]{ \inf_{g \in \spangl} \norm[\mu]{g - \varphi_i}^2} \geq   1 - \frac{r \cdot \paren{ 1 + \kappa(\Phi) }}{N}.
   \end{equation}
Therefore, there exists some $i \in [N]$ such that
 \[\EE[\bgi{1},\dotsc,\bgi{r}]{ \inf_{g \in \spangl} \norm[\mu]{g - \varphi_i}^2} \geq   1 - \frac{r \cdot \paren{ 1 + \kappa(\Phi) }}{N},\]
which gives us inequality \eqref{eq:randict1}. Inequality \eqref{eq:randict2} follows by an application of  Markov's inequality to the random variable $1 - \inf_{g \in \spangl} \norml[\mu]{g - \varphi_i}^2$ (which is easily seen to be non-negative), which by the first part of the theorem has expected value at most $\fracl{r \cdot \parenl{ 1 + \kappa(\Phi) }}{N}$.
 
 We conclude by proving the stronger version of the theorem, where we additionally assume that the random features are symmetric.
 Suppose $\bgi{1},\dotsc,\bgi{r}$ are symmetric with respect to $\Phi$.
    As mentioned in Fact \ref{fact:hilbert-proj}, the orthogonal projection $\Pi_{\spangl} \varphi_1$ depends on $\varphi_1$ only through the (random) vector $(\ipmeasl{\bgi{1}}{\varphi_1}{\mu}, \dotsc,\ipmeasl{\bgi{r}}{\varphi_1}{\mu})$.
  Therefore, by the symmetry assumption on the distribution of each $\bgi{i}$, the orthogonal projection $\Pi_{\spangl} \varphi_1$ has the same distribution as $\Pi_{\spangl} \varphi_i$ for all $i \in [N]$.
  Then
  \begin{align*}
    \EE[\bgi{1},\dotsc,\bgi{r}]{ \norm[\mu]{\Pi_{\spangl} \varphi_1}^2 }
    & = \frac1N \sum_{i=1}^N \EE[\bgi{1},\dotsc,\bgi{r}]{ \norm[\mu]{\Pi_{\spangl} \varphi_i}^2 }\stepcounter{equation}\tag{\theequation}\label{line:r}. \\
  \end{align*}
  Plugging Equation~\eqref{line:r} into Inequality~\eqref{eq:randict3} proves that Inequalities~\eqref{eq:randict1} and~\eqref{eq:randict2} hold for $i = 1$.
  
 \end{proof}

\subsection{MinWidth lower-bounds for RBL ReLU networks}\label{ssec:lb-relu-app}
Here, we specialize Theorem \ref{thm:randict} to the case of ReLU networks, which prepares us to prove the specific lower-bounds that will be given in the subsequent sections.

\begin{lemma}\label{lemma:randict-relu}
Let $\mathcal{D}$ be a symmetric ReLU parameter distribution and $\mu$ be some measure over $\R^d$. 
Fix any $\Phi = \setl{\varphi_1,\dotsc,\varphi_N} \subset \fnmeas{\mu}$ such that $\norml[\mu]{\varphi_i}^2 = 1$ for all $i \in [N]$.
Then, for any $\eps > 0$, there exists some $\varphi_i \in \Phi$ such that
\begin{equation}\label{eq:randict-relu1}
\minwidth{4 \eps \varphi_{i}}{\eps}{\half}{\mu}{\mathcal{D}} \geq \frac{N}{4 + 4\kappa(\Phi)} .
\end{equation}
Additionally, suppose that the functions in $\Phi$ are symmetric up to some permutation of variables and $\mu$ is invariant to permutation of variables.
That is, for all $i,i' \in [N]$ there exists a permutation $\pi_{i,i'}$ over $[d]$ such that $\varphi_i \circ \pi_{i,i'} = \varphi_{i'}$.
Then, Inequality \eqref{eq:randict-relu1} always holds for $i = 1$.
\end{lemma}
\begin{proof}		
	By applying Theorem \ref{thm:randict} for any $r \leq \fracl{N}{\parenl{4 + 4\kappa(\Phi)}}$ and for $\alpha = \fracl{1}{3}$, there exists some $i \in [N]$ such that 
	\[\pr[\bgi{1}, \dots, \bgi{r}]{\inf_{g \in \spangl} \norm[\mu]{\varphi_i - g} < \frac{1}{4}} < \frac{1}{2}.\]	
	Note that for all $f$, there exists $g \in \spangl$ with $\norml[\mu]{f - g} < \eps$ if and only if there exists $g' \in \spangl$ with $\norml[\mu]{\fracl{f}{4\eps}  - g'} < \fracl{1}{4}$. Thus, we conclude the following:
	\begin{align*}
		 \pr[\bgi{1}, \dots, \bgi{r}]{\inf_{g \in \spangl} \norm[\mu]{4\eps \varphi_i - g} < \eps} 
		&=  \pr[\bgi{1}, \dots, \bgi{r}]{\inf_{ g' \in \spangl} \norm[\mu]{\varphi_i - g'} < \frac{1}{4}} 
		< \half.
	\end{align*}
	
	To prove the stronger version of the theorem that assumes permutation symmetry for $\Phi$, we apply the stronger version of Theorem \ref{thm:randict}.
	To do so, we must show that each $\bgi{i}$ is symmetric with respect to $\Phi$. 
	 
	Because the ReLU feature parameters $\bbi{i}$ are chosen independently $\bwi{i}$ and the distribution of $\bwi{i}$ is invariant to variable permutation, each $\bgi{i}$ is drawn from a distribution that is also invariant to permutation. 
	We prove the symmetry property by showing that the inner product distributions are identical for $\bgi{1}$, without loss of generality.
	Because each function in $\varphi_1, \dots, \varphi_N$ is symmetric to a permutation of variables, there exists some permutation $\pi_{i, i'}$ such that for all $x \in \mu$, $\varphi_i(x) = \varphi_{i'}(\pi_{i,i'}(x))$. 
	To show that the two inner products induce the same distribution, consider any $z \in \R$.
  Then:
	\begin{align*}
    \lefteqn{
      \prl[\bgi{1}]{\ipmeasl{\bgi{1}}{\varphi_i}{\mu} \geq z}
    } \\
		&= \pr[\bgi{1}]{\EEl[\bx \sim\mu]{\bgi{1} (\bx) \varphi_i(\bx)} \geq z} \\
		&= \pr[\bgi{1}]{\EEl[\bx \sim \mu]{\bgi{1} (\bx) \varphi_j( \pi_{i,i'}(\bx))} \geq z} && \text{(Existence of $\pi_{i,i'}$)}  \\
		&= \pr[\bgi{1} ]{\EEl[\bx \sim \mu]{\bgi{1} (\pi_{i,i'}(\bx)) \varphi_j( \pi_{i,i'}(\bx))} \geq z} && \text{(Symmetry of $\bgi{1}$'s distribution)} \\
		&= \pr[\bgi{1}]{\EEl[\bx \sim \mu]{\bgi{1} (\bx) \varphi_{i'}(\bx)} \geq z} && \text{(Symmetry of $\mu$)}\\
		&= \prl[\bgi{1}]{\ipmeasl{\bgi{1}}{\varphi_{i'}}{\mu} \geq z}
	\end{align*}
	
	Hence, recalling Definition \ref{def:sym-random-fn}, $\bgi{1}$ is symmetric with respect to $\varphi_1,\dots,\varphi_N.$ 
	By invoking Theorem \ref{thm:randict} with the additional symmetry assumption, inequality \eqref{eq:randict-relu1} holds when $i = 1$.
\end{proof}

\subsection{Asymptotically tight lower-bounds for RBL ReLU networks over $\intvl^d$}\label{ssec:lb-relu-cube-tight-app}

To finalize the proof of Theorem \ref{thm:lb-cube-nonexplicit}, we first show that some low-degree trigonometric polynomial cannot be approximated by a combination of random ReLU features.\footnote{
We prove Lemma \ref{lemma:lb-cube-nonexplicit} separately from Theorem \ref{thm:lb-cube-nonexplicit} since we also make use of Lemma \ref{lemma:lb-cube-nonexplicit} in Appendix \ref{asec:lb-cube-sobolev} when proving lower-bounds based on the Sobolev norm of a function, rather than its Lipschitz constant.}

\begin{lemma}\label{lemma:lb-cube-nonexplicit}
	For any $k > 0$, any $\eps > 0$, and any symmetric ReLU parameter distribution $\mathcal{D}$, there exists some $K \in \N^d$ with $\norm[2]{K} \leq k$ such that
	\[\minwidth{4 \eps T_K}{\eps}{\half}{\intvl^d}{\mathcal{D}} \geq \frac{1}{4} Q_{k,d}.\] 
\end{lemma}
\begin{proofnoqed}
	Let $\mathcal{T}_k := \setl{T_K \in \mathcal{T}: K \in \mathcal{K}_{k,d}}$ be a subset of trigonometric basis elements with bounded degree.
	Because $\mathcal{T}$ is an orthonormal family of functions, $\mathcal{T}_k$ is as well, and $\kappa(\mathcal{T}_k) = 0$.
	Then, Lemma~\ref{lemma:randict-relu} implies the existence of some $T_K \in \mathcal{T}_k$ such that
	\[\minwidth{4 \eps T_K}{\eps}{\half}{\intvl^d}{\mathcal{D}} \geq \frac{\abs{\mathcal{T}_k}}{4} = \frac{1}{4} Q_{k,d}. \qedhere \]
\end{proofnoqed}
We prove Theorem \ref{thm:lb-cube-nonexplicit} by applying Lemma \ref{lemma:lb-cube-nonexplicit} and bounding the Lipschitz constant of the inapproximable function.

\begin{proof}[Proof of Theorem \ref{thm:lb-cube-nonexplicit}]
	Consider any $T_{K} \in \mathcal{T}$ with $\norm[2]{K} \leq k$.
	Then, for all $x, x' \in \intvl^d$,
	\[\abs{T_K(x) - T_K(x')} \leq \sqrt{2} \pi \innerprodl{K}{ x - x'} \leq \sqrt{2}\pi \norm[2]{K} \norml[2]{x - x'} \leq \sqrt{2} \pi k \norml[2]{x - x'}.\]
	
	Thus, $\norml[\lip]{T_K} \leq \sqrt{2} \pi k$ and $\norml[\lip]{f} \leq 4 \sqrt{2} \pi k \eps \leq 18 k\eps$.
	By applying Lemma \ref{lemma:lb-cube-nonexplicit} with $k := \fracl{L}{18 \eps},$ there exists a satisfactory $f$ such that $\norml[\lip]{f} \leq L$.
	\end{proof}

\subsection{Explicit lower-bounds for RBL ReLU networks over $\intvl^d$}\label{ssec:lb-relu-cube-explicit-app}

As in the previous section, we prove Lemma \ref{lemma:lb-cube-sine} by applying Lemma \ref{lemma:randict-relu} to a family of orthonormal functions. 
In order to obtain an explicit function $f$ that is hard to approximate, we invoke the stronger version of Lemma \ref{lemma:randict-relu}, which requires showing that that the family of functions exhibits symmetry up to a permutation of variables.

\begin{lemma}\label{lemma:lb-cube-sine}
	For any $\ell \in \Z^+$ with $\ell \leq d$, any $\eps > 0$, and any symmetric ReLU parameter distribution $\mathcal{D}$, define $f: \R^d \to \R$ to be the function $f(x) := 4 \sqrt{2} \eps \sin\parenl{\pi \sum_{i=1}^{\ell} x_i}$. Then,
	\[\minwidth{f}{\eps}{\half}{\intvl^d}{\mathcal{D}} \geq \frac{1}{4} {d \choose \ell}.\] %
\end{lemma}
\begin{proofnoqed}
	We prove the claim by constructing a family of functions $\Phi_{\ell}$ with $\frac{1}{4 \eps} f \in \Phi_{\ell}$ and applying Lemma \ref{lemma:randict-relu}.
	We define a family of functions 
	\[\Phi_{\ell} := \set{\varphi_S: x \mapsto \sqrt{2} \sin\paren{\pi \sum_{i\in S} x_i} \ \mid \ S \subseteq [d], \abs{S} = \ell}.\]
	Note that $\absl{\Phi_{\ell}} = {d \choose \ell}$ and that $\varphi_1 := \frac{1}{4 \eps}  f = \varphi_{[\ell]}\in \Phi_{\ell}$. 
	 Because $\Phi_{\ell} \subseteq \mathcal{T}$ and $\mathcal{T}$ is an orthonormal basis for $\fnmeas{\intvl^d}$ (Fact~\ref{fact:trig-orthonormal-basis-cube}), the functions in $\Phi_{\ell}$ are orthonormal and $\kappa(\Phi_{\ell}) = 0$. 
	Thus, because the $\Phi_{\ell}$ satisfies the symmetry conditions for the special case of Lemma~\ref{lemma:randict-relu},
	\[\minwidth{f}{\eps}{\half}{\intvl^d}{\mathcal{D}} \geq \frac{1}{4} {d \choose \ell}. \qedhere \]
\end{proofnoqed}

\begin{proofnoqed}[Proof of Theorem \ref{thm:lb-cube-sine}]
	This is immediate from Lemma \ref{lemma:lb-cube-sine} and from the fact that $\norm[\lip]{f} = 4  \pi \eps \sqrt{2\ell} \leq L$. 
	The right-hand side of the bound follows by lower-bounding ${d \choose \ell}$ for our choice of $\ell$.
	
	If $\ell = \ceill{\fracl{d}{2}}$ and $ d \geq 2$,\footnote{There is no need to consider the $d=1$ case, because then $\minwidth{f}{\eps}{\half}{\intvl^d}{\mathcal{D}} \geq \frac{1}{4} = \exp(\Theta(1))$, which satisfies the claim.} then
	\[{d \choose \ell} \geq \paren{\frac{d}{\ceil{\fracl{d}{2}}}}^{\ceil{\fracl{d}{2}}} \geq \paren{\frac{3}{2}}^{\fracl{d}{2}} \geq \exp\paren{\Theta(d)}.\]
	
	Otherwise, $\ell < \fracl{d}{2}$ and
	\[{d \choose \ell} \geq \paren{\frac{d}{\ell}}^{\ell} \geq \exp\paren{\Theta\paren{\ell \log\paren{\frac{d}{\ell}+2}}} = \exp\paren{\Theta\paren{\frac{L^2}{\eps^2}\log\paren{\frac{d\eps^2}{L^2}+2}}}. \qedhere \]
\end{proofnoqed}

This matches the exponent asymptotically up to logarithmic factors of the corresponding Lipschitz upper-bound, Theorem \ref{thm:ub-cube-lipschitz}.

\section{Upper- and lower-bounds for Sobolev functions}\label{asec:sobolev}

In this section, we present upper- and lower-bounds on the width required for depth-2 RBL ReLU approximation of functions in a larger family of smooth functions, namely the order-$s$ Sobolev functions. 
Sobolev spaces are normed function spaces arising in the study of partial differential equations, and their norms quantify the effective ``bumpiness'' of their constituent functions in terms of their weak derivatives.
Let $\mu$ denote the uniform probability measure on an open subset of $\R^d$.
Following \citet{leoni2017first}, we denote the \textit{order-$s$ Sobolev space of functions in $\fnmeas{\mu}$} for $s \in \N$ by\footnote{%
  Technically, $\dderiv{M}{f}$ is interpreted as the $M$-th weak partial derivative of $f$.
  However, it satisfies the integration-by-parts formulas that appear in the proof of Lemma~\ref{lemma:trig-expansion-deriv}, which is all we require.%
}
\[ \sobolev{s}{\mu} := \set{f : \R^d \to \R : \ \dderiv{M}{f} \in \fnmeas{\mu}, \ \forall M \in \N^d \text{ s.t. } |M| \leq s} . \]
The norm on this space is
\[ \norm[\sobolev{s}{\mu}]{f} := \sqrt{\sum_{|M| \leq s} \norm[\mu]{\dderiv{M}{f}}^2} . \]
(We do not consider Sobolev spaces in $L_p(\mu)$ for $p\neq2$ since we rely on Hilbert space structure.)

We focus on the classical spaces $\sobolev{s}{\mu}$ in $\fnmeas{\mu}$, where $\mu$ is the uniform product probability measure on the torus $\Circle^d$ and $\Circle = \R/(2\Z)$.
As a short-hand, we refer to this space as $\sobolev{s}{\Circle^d}$ in $\fnmeas{\Circle^d}$.
Recall that $\Circle$ is obtained by identifying points in $\R$ that differ by $2z$ for some $z \in \Z$.
Functions on $\Circle^d$ can be regarded as functions on $\intvl^d$, which, along with their derivatives, satisfy the periodic boundary conditions. 
Note that $\mathcal{T}$ is also an orthonormal basis for $\Circle^d$, because all of the trigonometric polynomials in $\mathcal{T}$ and all their derivatives have periodic boundary conditions and because the probability density of the uniform distribution on $\Circle^d$ is the same as the density over the uniform distribution on $\intvl^d$.

\subsection{Upper-bounds for functions in $\sobolev{s}{\Circle^d}$}\label{asec:ub-cube-sobolev}
We prove an analogue to Theorem~\ref{thm:ub-cube-lipschitz} that places an upper-bound on the minimum width RBL ReLU network that approximates a function with bounded order-$s$ Sobolev norm. 

\begin{theorem}\label{thm:ub-cube-sobolev}
	Fix some $\delta \in (0, \fracl{1}{2}]$, $\eps, \gamma > 0$, and $s \in \Z^+$. Let $k := \fracl{\sqrt{s} \gamma^{1/s}}{\eps^{1/s}}$. Then, there exists some ReLU parameter distribution $\mathcal{D}$ such that for any fixed $f \in \sobolev{s}{\Circle^d}$ that satisfies $\sobnorm{s}{\Circle^d}{f} \leq \gamma$, we have

	\[\minwidth{f}{\eps}{\delta}{\Circle^d}{\mathcal{D}} \leq O\paren{\frac{s^2 \gamma^{2 + 4/s}  d^2 }{\eps^{2 + 4/s}} Q_{k,d}^2 \ln \paren{\frac{1}{\delta}}}.\]
\end{theorem}
\begin{remark}
	When $s=1$,  
	\[\minwidth{f}{\eps}{\delta}{\Circle^d}{\mathcal{D}} \leq O\paren{\frac{ \gamma^{6}  d^2 }{\eps^{6}} Q_{\gamma / \eps,d}^2 \ln \paren{\frac{1}{\delta}}},\]
	which is a near-perfect match to the upper-bound for Lipschitz functions in Theorem~\ref{thm:ub-cube-lipschitz}.
	This is unsurprising, because all $L$-Lipschitz functions $f$ with $\abs{\EE{f}} \leq L$ have a squared 1-order Sobolev norm with the following bound:
	\[\sobnorm{s}{\Circle^d}{f}^2 = \norm[\Circle]{f}^2 + \EE[\bx \sim \Circle^d]{\norm{\nabla f(\bx)}^2} \leq O(L^2).\]
	Thus, the two theorems give nearly identical upper-bounds for $L$-Lipschitz functions that satisfy periodic boundary conditions.
\end{remark}
\begin{remark}\label{rem:sob-ub}
	Applying Fact~\ref{fact:Qkd-ub} to Theorem~\ref{thm:ub-cube-sobolev} implies that 
	\[\minwidth{f}{\eps}{\delta}{\Circle^d}{\mathcal{D}} \leq  \ln \paren{\frac{1}{\delta} }\exp\paren{O\paren{\min\paren{d \log\paren{\frac{s \gamma^{2/s}}{d \eps^{2/s}}+2}, \frac{s\gamma^{2/s}}{\eps^{2/s}} \log\paren{\frac{d\eps^{2/s}}{s \gamma^{2/s}}+2}}}}.\]
\end{remark}

Like the proof of Theorem~\ref{thm:ub-cube-lipschitz}, we first show that every function in $\sobolev{s}{\Circle^d}$ can be approximated by low-degree trigonometric polynomial in Lemma \ref{lemma:trig-sobolev-approx}, which is a parallel result to Lemma \ref{lemma:trig-lipschitz-approx}.
Unlike Theorem~\ref{thm:ub-cube-lipschitz}, however, we require that $f$ and its first $s$ derivatives satisfy the periodic boundary conditions, which is assured by the fact that $f \in \sobolev{s}{\Circle^d}$.
Thanks to this assumption, we eliminate the need for the ``reflection'' trick from Lemma \ref{lemma:trig-lipschitz-approx}, which  simplifies the proof. 

\begin{lemma}[Approximating Sobolev functions with low-degree trigonometric polynomials]\label{lemma:trig-sobolev-approx}
	Fix any values $\gamma, \eps > 0$ and $s \in \mathbb{Z}^+$.
  Consider any $f \in \sobolev{s}{\Circle^d}$ with $\sobnorm{s}{\Circle^d}{f} \leq \gamma$.
	Let $k := \fracl{\sqrt{s} \gamma^{1/s}}{ (2\eps)^{1/s}}$.
  Then, there exists a trigonometric polynomial 
\[P(x) = \displaystyle\sum_{K \in \mathcal{K}_{k,d}} \beta_K T_K(x)\]
  such that 
	$ \norm[\Circle^d]{f - P} \leq \eps . $
  Moreover, $\abs{\beta_K} \leq \norm[\Circle^d]{f} \leq \gamma$ for all $K \in \mathcal{K}_{k,d}$.
\end{lemma}

\begin{proof}
	Because $\mathcal{T}$ is an orthonormal basis over $\Circle^d$, we express $f$ as the expansion
	\[ f = \sum_{K \in \Z^d} \alpha_K T_K . \]
  Since $f$ can be regarded as a function on $\intvl^d$ whose first $s$ partial derivatives satisfy boundary conditions, Lemma~\ref{lemma:trig-expansion-deriv} implies that this expansion of $f$ can be differentiated term-by-term.
  By taking term-by-term partial derivatives of $f$, applying Parseval's identity (Fact~\ref{fact:parseval-plancherel}), and using the known norms of partial derivatives of  $T_K$ (Fact~\ref{fact:trig-pderiv}), we obtain the following closed-form $\fnmeas{\Circle^d}$ norm for $\dderiv{M}{f}$ for all $M \in \N^d$ with $\abs{M} \leq s$:
  \[
    \norm[\Circle^d]{\dderiv{M}{f}}^2 = \sum_{K \in \Z^d} \alpha_K^2 (\pi K)^{2M} .
  \]
  Therefore, the squared $\sobolev{s}{\Circle^d}$-norm of $f$ can be written as  \begin{equation}\label{eq:sobolev-norm-expansion}
  \sobnorm{s}{\Circle^d}{f}^2  = \sum_{|M|\leq s} \norm[\Circle^d]{\dderiv{M}{f}}^2  = \sum_{|M|\leq s} \sum_{K \in \Z^d} \alpha_K^2 (\pi K)^{2M} = \sum_{K \in \Z^d} \alpha_K^2 c_{K,s},
 \end{equation}
 where
 \[c_{K,s} := \sum_{ |M| \leq s} (\pi K)^{2M}.\]
 We lower-bound $c_{K,s}$ in terms of $s$ and $\norm[2]{K}$ with the multinomial theorem:
 \begin{align*}
 	c_{K,s} &\geq \sum_{\substack{ |M| = s}} (\pi K)^{2M} \geq \frac{\pi^{2s}}{s!} \sum_{\substack{|M| = s}} \frac{s!}{M!} K^{2M} = \frac{\pi^{2s}}{s!} \norm[2]{K}^{2s} \geq \paren{\frac{\pi^2 \norm[2]{K}^2}{s}}^s.
 \end{align*}
We define $\beta_K := \alpha_K$ for all $K \in \mathcal{K}_{k,d}$ and $\beta_K := 0$ for all other $K \in \Z^d$.
Note that if $K \in \Z^d$ has $\norm[2]{K} > k \geq  \fracl{\sqrt{s} \gamma^{1/s}}{\pi \eps^{1/s}}$, then $c_{K,s} > \fracl{\gamma^2}{\eps^2}$.
  By Parseval's identity, we have $\beta_K^2 \leq \norm[\Circle^d]{f}^2$.
  Moreover,
  \begin{equation*}
    \norm[\Circle^d]{f - P}^2
    = \sum_{K \in \Z^d \setminus \mathcal{K}_{k,d}} \alpha_K^2
    \leq \sum_{\substack{K \in \Z^d : \\ c_{K,s} > \gamma^2/\eps^2}} \alpha_K^2
    \leq \sum_{\substack{K \in \Z^d : \\ c_{K,s} > \gamma^2/\eps^2}} \alpha_K^2 \cdot \frac{c_{K,S}}{\gamma^2 / \eps^2}
    \leq \frac{\eps^2}{\gamma^2} \sum_{K \in \Z^d} \alpha_K^2 c_{K,s}
    \leq \epsilon^2 .
  \end{equation*} 
  Above, the first equality uses Parseval's identity, and the final equality uses Equation~\eqref{eq:sobolev-norm-expansion}.
\end{proof}

\begin{proof}[Proof of Theorem \ref{thm:ub-cube-sobolev}]
	This proof is identical to the proof of Theorem~\ref{thm:ub-cube-lipschitz} in Section~\ref{ssec:ub-cube-lipschitz-proof}, except that we make use of Lemma \ref{lemma:trig-sobolev-approx} instead of Lemma \ref{lemma:trig-lipschitz-approx}, and instead set $k :=  \fracl{\sqrt{s} \gamma^{1/s}}{\eps^{1/s}}$ and $\rho := 1$.
\end{proof}

\subsection{Lower-bounds for functions in $\sobolev{s}{\intvl^d}$}\label{asec:lb-cube-sobolev}
Similar to Section~\ref{sec:lb}, we give lower-bounds on the width of RBL ReLU neural networks required to approximate certain functions (now ones with bounded $s$-order Sobolev norm).
As before, we present two variants of the lower-bound, one non-explicit tight bound and one looser explicit bound.
\begin{itemize}
	\item Theorem~\ref{thm:lb-cube-sobolev} is analogous to Theorem~\ref{thm:lb-cube-nonexplicit}. 
	It shows the existence of some sinusoidal function with bounded Sobolev norm which matches the upper-bound Theorem~\ref{thm:ub-cube-sobolev} by depending on the same combinatorial term.
	\item Theorem~\ref{thm:lb-cube-sobolev-explicit}, like Theorem~\ref{thm:lb-cube-sine}, offers an explicit sinusoidal function with bounded Sobolev norm whose minimum width can be bounded by a term that differs from the asymptotics of the exponent of the upper-bound by a logarithmic factor. 
\end{itemize}

These results follow from proofs that directly apply Lemmas~\ref{lemma:lb-cube-nonexplicit} and~\ref{lemma:lb-cube-sine} respectively and bound the $s$-order Sobolev norms of the resulting functions.

\subsubsection{A tight lower-bound}\label{sssec:lb-tight}
We give a bound on the minimum width depth-2 RBL ReLU network needed to approximate some function with bounded Sobolev norm, which is a scaled version of some function in $\mathcal{T}$.
The family of functions is identical to that of Theorem~\ref{thm:lb-cube-sobolev}; the only difference is that we parameterize the bounds by the $s$-order Sobolev norm of the function, rather than its Lipschitz constant.

\begin{theorem}\label{thm:lb-cube-sobolev}
	Fix some $\eps, \gamma> 0$ and $s \in \Z_+$ with $\fracl{\gamma^2}{\eps^2} \geq 16(s+1)$. Let \[k:= \frac{\gamma^{1/s}}{\pi 4^{1/s} \eps^{1/s}(s+1)^{1/2s}}.\]
	Then, there exists some $K \in \mathcal{K}_{k,d}$ such that for $f := 4\eps T_{K}$ and for any symmetric ReLU parameter distribution $\dprod$,

	\[\minwidth{f}{\eps}{\half}{\Circle^d}{\mathcal{D}} \geq  \frac{1}{4} Q_{k,d},\]
	and $\sobnorm{s}{\Circle^d}{f} \leq \gamma$.	
\end{theorem}

\begin{remark}
	By invoking Fact~\ref{fact:Qkd-ub}, we have 
	\[\minwidth{f}{\eps}{1/2}{\Circle^d}{\mathcal{D}} \geq  \exp\paren{\Omega\paren{\min\paren{d \log\paren{\frac{\gamma^{2/s}}{d \eps^{2/s}}+2}, \frac{\gamma^{2/s}}{\eps^{2/s}} \log\paren{\frac{d\eps^{2/s}}{ \gamma^{2/s}}+2}}}}.\]
	Note that we can drop $(s+1)^{1/s}$ terms from the asymptotics of the exponent, because $(s+1)^{1/s} \in (1, 2]$ for all $s \in \Z^+$.
	The asymptotics of the exponents match the upper-bound on the minimum width presented in Remark~\ref{rem:sob-ub}, when $\delta = \fracl{1}{2}$ and $s$ is regarded as a small constant.
\end{remark}

\begin{proof}
	To prove the existence of $f$, we need only invoke Lemma~\ref{lemma:lb-cube-nonexplicit} for our choice of $k$. 
	It remains to bound the $s$-order Sobolev norm of $f$. 
	We do so by expanding the squared Sobolev norm of $f$ and applying Fact~\ref{fact:trig-pderiv} to obtain an exact representation of the norms of derivatives of the basis elements $T_K \in \mathcal{T}$. 
	\begin{align*}
		\sobnorm{s}{\Circle^d}{f}^2
		&= \sum_{M: \abs{M} \leq s} \norm[\Circle^d]{\dderiv{M}{f}}^2
		= 16 \eps^2 \sum_{M: \abs{M} \leq s} \norm[\Circle^d]{\dderiv{M}{T_K}}^2 
		= 16 \eps^2 \sum_{M: \abs{M} \leq s} \pi^{2\abs{M}} K^{2M} \\
		&= 16 \eps^2 \sum_{m=0}^s \pi^{2m} \sum_{\abs{M} = m} K^{2M}
		\leq 16 \eps^2 \sum_{m=0}^s \pi^{2m} \sum_{\abs{M} = m} \frac{m!}{K!} K^{2M}
		= 16 \eps^2 \sum_{m=0}^s \pi^{2m} \norm[2]{K}^{2m} \\
		&\leq 16 \eps^2 \sum_{m=0}^s \paren{\pi^2 k^2}^m
		= 16 \eps^2 \sum_{m=0}^s \paren{\frac{\gamma^{2/s}}{16^{1/s} \eps^{2/s}(s+1)^{1/s}}}^m
	\end{align*}
	Because of our assumed lower-bound on $\fracl{\gamma^2}{\eps^{2}}$, the final term of the sum cannot be smaller than any preceding terms.  
	Therefore, we conclude with the following trivial bound on the sum.
	\begin{align*}
		\sobnorm{s}{\Circle^d}{f}^2
		&\leq 16 \eps^2 \sum_{m=0}^s \paren{\frac{\gamma^{2/s}}{16^{1/s} \eps^{2/s}(s+1)^{1/s}}}^m
		\leq16 \eps^2 (s+1) \paren{\frac{\gamma^{2/s}}{16^{1/s} \eps^{2/s}(s+1)^{1/s}}}^s
		= \gamma^2. 
	\end{align*}
\end{proof}

\subsubsection{A lower-bound for an explicit sinusoidal function}\label{sssec:lb-explicit}
We give an explicit lower-bound that bounds the Sobolev norm of the function $f$ used in Lemma \ref{lemma:lb-cube-sine}.
In that way, it is nearly identical to Theorem~\ref{thm:lb-cube-sine}.

\begin{theorem}\label{thm:lb-cube-sobolev-explicit}
	Fix some $\eps, \gamma> 0$ and $s \in \Z_+$ with $\fracl{\gamma^2}{\eps^2} \geq 16(s+1)$.
	 Let \[\ell := \min\paren{\ceil{\frac{d}{2}}, \floor{ \frac{\gamma^{2/s}}{\pi^2 16^{1/s} \eps^{2/s}(s+1)^{1/s}}}}.\]
 Fix any symmetric ReLU parameter distribution $\dprod$.
 Then, the function $f(x) := 4 \sqrt{2} \eps \sin\parenl{\pi \sum_{i=1}^{\ell} x_i}$ satisfies $\sobnorm{s}{\Circle^d}{f} \leq \gamma$ and 
	\[\minwidth{f}{\eps}{\half}{\Circle^d}{\mathcal{D}} \geq \frac{1}{4} {d \choose \ell} \geq \exp\paren{\Omega\paren{\min\paren{\frac{\gamma^{2/s}}{\eps^{2/s}} \log\paren{\frac{d\eps^{2/s}}{\gamma^{2/s}}+2}, d}}}.\]
\end{theorem}

\begin{proof}
	The width bound is immediate from Lemma \ref{lemma:lb-cube-sine} and from the lower-bounds on ${d \choose \ell}$ shown in the proof of Theorem \ref{thm:lb-cube-sine}. 
	Note that $f$ can be written as $f = 4\eps T_K$ for some $K$ with 
	\[\norm[2]{K} = \sqrt{\ell}\leq  \frac{\gamma^{1/s}}{\pi 4^{1/s} \eps^{1/s}(s+1)^{1/2s}}.\]
	Thus, we conclude that $\sobnorm{s}{\Circle^d}{f} \leq \gamma$ by applying the same chain of inequalities from Theorem~\ref{thm:lb-cube-sobolev}, making use of our lower-bound on $\fracl{\gamma^2}{\eps^2}$.
\end{proof}

%

\section{A similar approach for the Gaussian measure}\label{asec-gaussian}

The techniques underlying our upper- and lower-bounds on approximation by depth-2 RBL networks are rather general, and can be applied in a broader range of settings than are captured by Theorems~\ref{thm:ub-cube-lipschitz-informal} and~\ref{thm:lb-cube-sine-informal}. These settings include other activation functions beyond ReLU gates  and other functions spaces beyond $\fnmeas{\intvl^d}$.
In this Appendix, we briefly sketch how several of the key ingredients for Theorems~\ref{thm:ub-cube-lipschitz-informal} and~\ref{thm:lb-cube-sine-informal} have analogues over \emph{Gaussian space}, and how results similar to Theorems~\ref{thm:ub-cube-lipschitz-informal} and~\ref{thm:lb-cube-sine-informal} can be proved over Gaussian space.\footnote{%
  Coarse analogues of the results from Appendix~\ref{asec:sobolev} for Sobolev spaces may also be obtained with these techniques.%
}

\subsection{The setting and key background results}

We consider the domain $\R^d$ endowed with the standard $d$-dimensional Gaussian measure $\MNormal$ with mean zero and identity covariance matrix. It is well known (see e.g.~Section 11.2 of \cite{odonnell14}) that the set $\{H_K\}_{K \in \N^d}$ of all \emph{multivariate normalized Hermite polynomials} is an orthonormal basis for $\fnmeas{\MNormal}$, where for $K=(K_1,\dots,K_d)$ the function $H_K$ is 
\[
H_K = \prod_{j=1}^d  h_{K_j}(x_j)
\]
where $h_i$ is the degree-$i$ normalized univariate Hermite polynomial.  These multivariate Hermite polynomials are analogous to the trigonometric basis polynomials $T_K$ that are introduced in Appendix~\ref{sec:trig-basis} for the function space $\fnmeas{\intvl^d}.$

Well known results (see, e.g., Section~5.5 of \cite{Szego:39}) show that partial derivatives of multivariate normalized Hermite polynomials can be conveniently expressed in terms of other multivariate normalized Hermite polynomials, very analogous to Equation~\ref{eq:deriv-of-trig}.  By combining this with a well-known recurrence relation for Hermite polynomials (again, see \cite{Szego:39}), it is possible to prove the following result, which is closely analogous to Lemma \ref{lemma:trig-expansion-deriv} but now for $\fnmeas{\MNormal}$ rather than $\fnmeas{\intvl^d}$:

\begin{lemma}[Term-by-term differentiation for Hermite representation]\label{lemma:hermite-expansion-deriv}
  \sloppy
	Consider some $f \in \fnmeas{\MNormal}$ and $i \in [d]$ such that $f$ is differentiable with respect to $x_i$ and $\pderiv{f}{x_i} \in  \fnmeas{\MNormal}$. Then, $f$ and its partial derivative $\pderivl{f}{x_i}$ have Hermite expansions of the form
\[
f = \sum_{K \in \N^d} \alpha_K H_K 
\quad \& \quad
\pderiv{f}{x_i} =  \sum_{K \in \N^d} \alpha_{K} \pderiv{H_{K}}{x_i}.
\]
  \fussy
\end{lemma}

\subsection{The upper-bound approach}

Recall that our positive results for depth-2 RBL ReLU approximation are proved in two stages.  In the first stage (Lemma~\ref{lemma:trig-lipschitz-approx}, restated as Lemma~\ref{lemma:trig-lipschitz-approx-restate} in Appendix~\ref{ssec:trig-lipschitz-approx}), we argued that any Lipschitz function over $\intvl^d$ can be approximated as a low-degree trigonometric polynomial with bounded coefficients.  In the second stage (Lemma~\ref{lemma:ub-cube-trig-bounded-degree}), we argued that low-degree trigonometric polynomials can be approximated with depth-2 RBL ReLU networks.

For the first stage, with Lemma~\ref{lemma:hermite-expansion-deriv} in hand as an analogue of Lemma \ref{lemma:trig-expansion-deriv}, it is possible to obtain an analogue of Lemma~\ref{lemma:trig-lipschitz-approx-restate}; in the current Gaussian setting, this result shows that functions in $\fnmeas{\MNormal}$ with bounded Lipschitz constant can be approximated with low-degree Hermite polynomials whose coefficients (in terms of the orthonormal basis of normalized multivariate Hermite polynomials) are not too large.  (The argument is in fact simpler than for Lemma~\ref{lemma:trig-lipschitz-approx-restate} because there are no issues with periodic boundary conditions, which were responsible for steps~1, 2, 4 and 5 of the outline provided at the beginning of the proof of Lemma~\ref{lemma:trig-lipschitz-approx-restate}.)

For the second stage, some technical challenges arise because the Hermite basis functions (unlike the trigonometric polynomials defined in Appendix~\ref{sec:trig-basis}) are not ridge functions.  These challenges can be overcome: using techniques from \cite{apvz}, it is possible to show that the small-coefficient, low-degree Hermite polynomials we are dealing with can indeed be approximated by depth-2 RBL ReLU networks. It turns out that the resulting width of the RBL ReLU networks obtained using this approach is roughly $(dL/\eps)^{O(L^2/\eps^2)}$, i.e., polynomial in the dimension $d$ but exponential in $L/\eps$; this corresponds to a somewhat weaker analogue of Theorem~\ref{thm:ub-cube-lipschitz} in which the ``$\min(L^2/\eps^2,d)$'' is replaced with just $L^2/\eps^2$, and gives a good upper bound when $d$ is large compared to $L^2/\eps^2$.  For the complementary regime where $L^2/\eps^2$ is large compared to $d$, using different techniques\footnote{Roughly speaking, the approach (inspired by \cite{jtx19}) is to (i) truncate the function by setting it to a constant outside of a ball of carefully chosen radius; (ii) approximate the truncated function with a superposition of ``Gaussian bumps;'' (iii) approximate this superposition of Gaussian bumps by a weighted average of random ReLU gates.} it can be shown that in fact depth-2 RBL ReLU networks of width roughly $(dL/\eps)^{O(d)}$ also suffice; combining these two regimes, this gives an overall approximation result for Lipschitz functions over Gaussian space that is  quite closely analogous to Theorem~\ref{thm:ub-cube-lipschitz}.  The arguments to establish these results are somewhat lengthy for each of the two regimes, though, so we omit both the arguments and detailed claims of the results in this paper. 
\clearpage
\subsection{The lower-bound approach}

Recall that our main lower bound tool, Theorem~\ref{thm:randict}, only requires small average coherence (rather than strict orthogonality) for the set of ``hard'' functions .  Exploiting this flexibility, it is not difficult to adapt Theorem~\ref{thm:randict} to the setting of Gaussian space.  

In a bit more detail, it turns out that taking a family $\Phi = \setl{\varphi_1,\dotsc,\varphi_N}$ of ``hard'' functions that corresponds to points $v^{(1)},\dots,v^{(N)}$ in a suitable packing of the unit sphere, where the function $\varphi_i(x)$ is defined to be (a suitably normalized version of) $\sin(L \innerprodl{v^{(i)}}{x})$, results in $\Phi$ having small average coherence, and from this it is not difficult to obtain lower-bounds on depth-2 RBL ReLU network width, following the approach of Section~\ref{sec:lb}.  The resulting lower bounds can be shown to be quite close to matching the upper-bounds for Gaussian space sketched in the previous subsection.

\end{document}